\documentclass[10pt,journal]{article}

\usepackage{epsfig}
\usepackage{url}

\usepackage{psfrag}
\usepackage{pst-all}
\usepackage{times}
\usepackage{epsfig}
\usepackage{graphicx}
\usepackage{amsmath}
\usepackage{amssymb}
\usepackage{algorithm}
\usepackage{algorithmic}
\usepackage{nameref}
\usepackage{subfigure}
\usepackage[numbers]{natbib}
\usepackage{fullpage}
\usepackage{graphicx}
\usepackage{nameref}
\usepackage{times}
\usepackage{epsfig}
\usepackage{amsmath}
\usepackage{amssymb}
\usepackage{algorithm}
\usepackage{algorithmic}
\usepackage{subfigure}
\usepackage{booktabs}
\usepackage[flushleft]{threeparttable}
\usepackage{natbib}
\setcitestyle{parenthesis}
\usepackage{hyperref}
\usepackage{url}

\begin{document}
%
\title{A Deep and Autoregressive Approach for \\Topic Modeling of Multimodal Data}

\author{Yin Zheng\\
Department of Electronic Engineering, \\Tsinghua University, \\Beijing, China, 10084\\
{\tt\small y-zheng09@mails.tsinghua.edu.cn}
\and
Yu-Jin Zhang\\
Department of Electronic Engineering\\ Tsinghua University,\\ Beijing, China, 10084\\
{\tt\small zhang-yj@tsinghua.edu.cn}
\and
Hugo Larochelle\\
D\'{e}partment d'Informatique\\ Universit\'{e} de Sherbrooke, Sherbrooke (QC), Canada, J1K 2R1\\
{\tt\small hugo.larochelle@usherbrooke.ca }
}
\maketitle

\begin{abstract}
Topic modeling based on latent Dirichlet allocation (LDA) has been a
  framework of choice to deal with multimodal data, such as 
  in image annotation tasks. Another popular approach to model the multimodal data is through deep neural networks, such as the deep Boltzmann machine (DBM). 
  Recently, a new type of topic model called the Document
  Neural Autoregressive Distribution Estimator (DocNADE) was
  proposed and demonstrated state-of-the-art performance for text document
  modeling. In this work, we show how to successfully apply and extend
  this model to multimodal data, such as simultaneous image classification and annotation.
  First, we propose SupDocNADE, a supervised extension of DocNADE, that
  increases the discriminative power of the learned hidden topic features
  and show how to employ it to learn a joint representation 
  from image visual words, annotation words and class label information. 
  We test our model on the LabelMe and UIUC-Sports
  data sets and show that it compares favorably to other
  topic models.
Second, we propose a deep extension of our model and provide an efficient way of training the deep model. Experimental results show that
our deep model outperforms its shallow version and reaches state-of-the-art performance on the
Multimedia Information Retrieval (MIR) Flickr data set.
\end{abstract}


\section{Introduction}
%
%

%
%
%
%
Multimodal data modeling, which combines information from 
different sources, is increasingly attracting attention in computer vision~\cite{barnard2003matching,blei2003modeling,socher2010connecting,jia2011learning,putthividhy2010topic, guillaumin2010multimodal,rasiwasia2010new}. 
One of the leading approaches is based on topic
modelling, the most popular model being latent Dirichlet allocation or
LDA~\cite{blei2003latent}. LDA is a generative model for documents
that originates from the natural language processing community, but
has had great success in computer vision~\cite{blei2003latent, wang2009simultaneous}. 
LDA models a document as a multinomial distribution over topics, where a topic is
itself a multinomial distribution over words.  While the distribution
over topics is specific for each document, the topic-dependent
distributions over words are shared across all documents. Topic models
can thus extract a meaningful, semantic representation from a document
by inferring its latent distribution over topics from the words it
contains. In the context of computer vision, LDA can be used by first
extracting so-called ``visual words'' from images, convert the images
into visual word documents and training an LDA topic model on the
bags of visual words. 
\begin{figure}[t]
\begin{center}
	\includegraphics[width=0.65\linewidth]{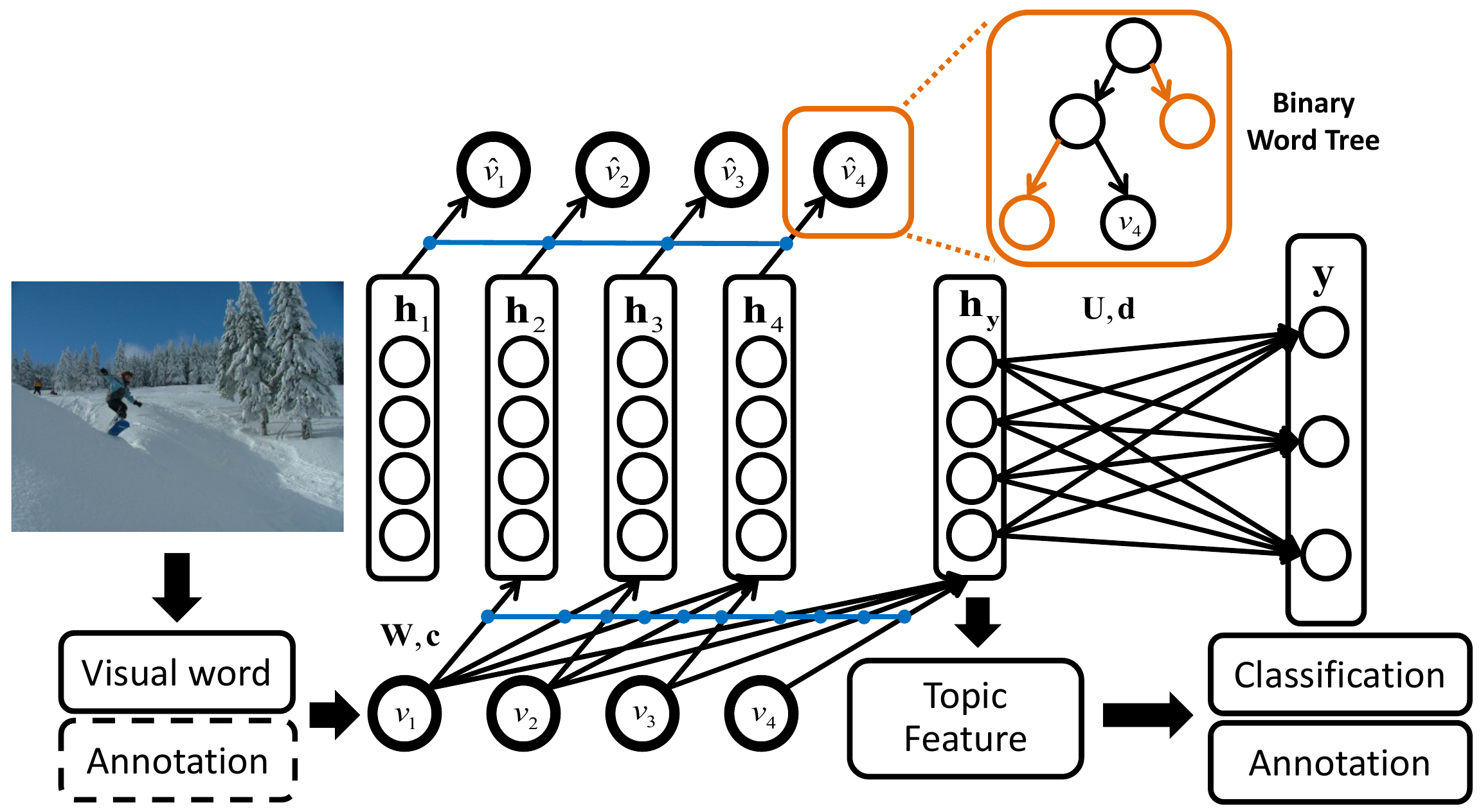}
\end{center}
\caption{Illustration of a single hidden layer SupDocNADE model for multimodal image data. 
  Visual words, annotation words and class 
  label $y$ are modeled as $p({\bf v},y) = p(y|{\bf v}) \prod_i
  p(v_i|v_1,\dots,v_{i-1})$.  All conditionals $p(y|{\bf v})$ and $
  p(v_i|v_1,\dots,v_{i-1})$ are modeled using neural networks with
  shared weights. Each predictive word conditional
  $p(v_i|v_1,\dots,v_{i-1})$ (noted ${\hat v}_i$ for brevity)
  follows a tree decomposition where each leaf is a possible word. At
  test time, the annotation words are not used (trated with a
  dotted box) to compute the image's topic feature representation.}
\label{fig:supdocnade}
\end{figure}
To deal with multimodal data, some variants of LDA
have been proposed recently~\cite{blei2003modeling, putthividhy2010topic, jia2011learning, wang2009simultaneous}.
 For instance, Correspondence LDA (Corr-LDA)~\cite{blei2003modeling}
was proposed to discover the relationship between images and annotation modalities, by assuming
each image topic must have a corresponding text topic. Multimodal LDA~\cite{putthividhy2010topic}
generalizes Corr-LDA by learning a regression module relating the topics from the different modalities.
Multimodal Document Random Field Model (MDRF)~\cite{jia2011learning} was also proposed
to deal with multimodal data, which learns cross-modality similarities from a document corpus
containing multinomial data.
Besides the annotation words, the class label modality can also be embedded into LDA, 
such as in supervised LDA (sLDA)~\cite{blei2007supervised, wang2009simultaneous}.
By modeling the image visual words, annotation words and their class labels, the
discriminative power of the learned image representations could thus be
improved.

At the heart of most topic models is a generative story in which the
image's latent representation is generated first and the visual words
are subsequently produced from this representation. The appeal of this
approach is that the task of extracting the representation from
observations is easily framed as a probabilistic inference problem,
for which many general purpose solutions exist.  The disadvantage
however is that as a model becomes more sophisticated, inference
becomes less trivial and more computationally expensive. In LDA for
instance, inference of the distribution over topics does not have a
closed-form solution and must be approximated, either using
variational approximate inference or MCMC sampling. Yet, the model is
actually relatively simple, making certain simplifying independence
assumptions such as the conditional independence of the visual words
given the image's latent distribution over topics.

Another approach to model the statistical structure of 
words is through the use of distributed representations modeled by
artificial neurons. In the realm of document modeling,
\citet{salakhutdinov2009replicated} proposed a so-called Replicated Softmax (RS) model for bags of words. The RS model was later used for 
multimodal data modeling~\cite{srivastava2012multimodal}, where pairs
of images and text annotations were modeled jointly within a deep Boltzmann
machine (DBM)~\cite{srivastava2013discriminative}. This deep learning approach to the generative modeling of multimodal data achieved state-of-the-art performance on the MIR Flickr data set~\cite{HuiskesM2008}.
On the other hand, it also shares with LDA and its different extensions the reliance on a stochastic latent representation of the data, requiring variational approximations and MCMC sampling at training and test time.
 Another neural network based state-of-the-art multimodal data modeling approach is Multimodal Deep Recurrent Neural Network (MDRNN)~\cite{sohn2014improved} which aims at predicting missing data modalities
through the rest of data modalities by minimizing the variation of information rather than maximizing likelihood.

Recently, an alternative generative modeling approach for documents
was proposed in \citet{larochelle2012neural}. In this work, a Document Neural
Autoregressive Distribution Estimator (DocNADE) is proposed, which models directly the
joint distribution of the words in a document by decomposing it
as a product of conditional distributions (through the probability chain rule) 
and modeling each conditional using a neural network. 
Hence, DocNADE doesn't incorporate any latent random
variables over which potentially expensive inference must be
performed. Instead, a document representation can be computed efficiently
in a simple feed-forward fashion, using the value of the neural
network's hidden layer. \citet{larochelle2012neural} also show that DocNADE is a better
generative model of text documents than LDA and the RS model, and can extract a useful 
representation for text information retrieval.

\begin{figure}[t]
\begin{center}
	\includegraphics[width=0.65\linewidth]{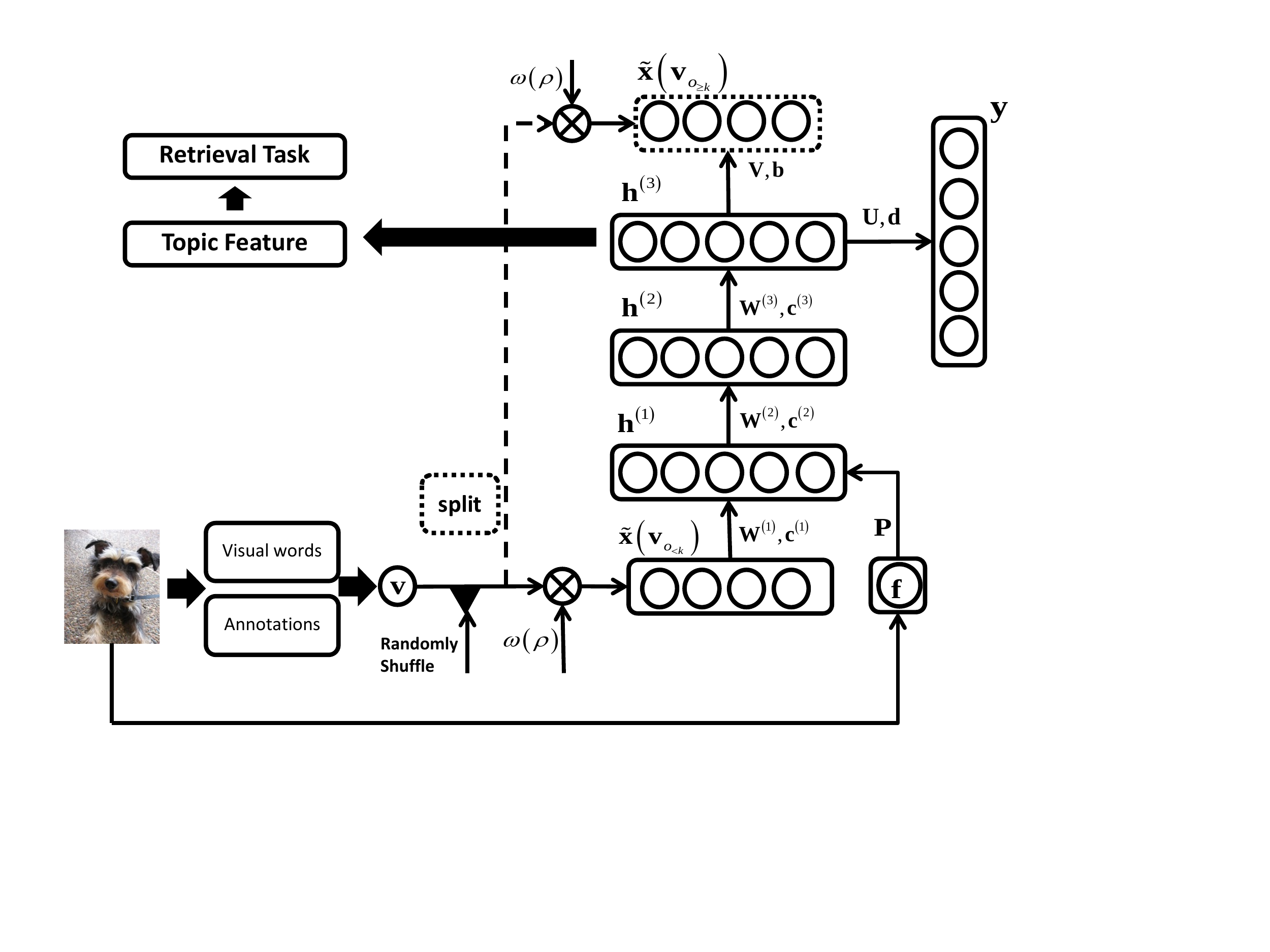}
\end{center}
\caption{Illustration of the deep extension of Supervised DocNADE (SupDeepDocNADE) model.  At the 
training phase, the input $\bf v$ (visual and annotation words) is first shuffled randomly based on an ordering $o$ and then randomly split into two parts, $\mathbf{v}_{o_{<d}}$ and $\mathbf{v}_{o_{\geq d}}$.  Then we compute each of the conditionals in Equation~\ref{eqn:DocNADE_est} and use backpropagation to optimize the  parameters of the model. 
To deal with the imbalance between the visual and annotation words, the histogram of $\mathbf{v}_{o_{<d}}$ and $\mathbf{v}_{o_{\geq d}}$ is weighted by $\omega \left(\rho\right)$. At test time, all the words in $\bf v$ are fed to the model to compute a discriminative deep representation. Besides the visual and annotation words, global features $\bf f$ are also leveraged by the model. }
\label{fig:supdeepdocnade}
\end{figure}

In this paper, we consider the application of DocNADE to deal with multimodal data 
in computer vision. More specifically, we first propose a supervised
variant of DocNADE (SupDocNADE), which can be used to model the joint distribution
over an image's visual words, annotation words and class label. The
model is illustrated in Figure~\ref{fig:supdocnade}. We investigate how to
successfully incorporate spatial information about the visual words
and highlight the importance of calibrating the generative and
discriminative components of the training objective. Our results
confirm that this approach can outperform other topic models, such as the supervised variant of LDA. Moreover, we propose a deep extension of SupDocNADE, that learns a deep and discriminative representation of pairs of images and annotation words. The deep version of SupDocNADE, which is illustrated in Figure~\ref{fig:supdeepdocnade}, outperforms its shallow one and achieves state-of-the-art performance on the challenging MIR Flickr data set.

\section{Related Work}
\label{relaged works}

As previously mentioned, multimodal data is often modeled
using extensions of the basic LDA topic model, such as Corr-LDA~\cite{blei2003modeling}, 
Multimodal LDA~\cite{putthividhy2010topic} and MDRF~\cite{jia2011learning}. 
In this paper, we focus on learning a joint representation from three different modalities: \textit{image
visual words, annotations}, and \textit{class labels}. The class label describes the
image globally with a single descriptive label (such as {\it coast},
{\it outdoor}, {\it inside city}, etc.), while the annotation focuses on tagging
the local content within the image.
\citet{wang2009simultaneous} proposed a supervised LDA formulation
to tackle this problem. \citet{wang2011max} opted instead for a
maximum margin formulation of LDA (MMLDA). Our work also belongs to
this line of work, extending topic models to a supervised variant:
 our first contribution in this paper is thus to extend a different topic model,
DocNADE, to this context for multimodal data modeling.

What distinguishes DocNADE from other topic models is its reliance on
an autoregressive neural network architecture. Recently, deep neural networks are increasingly used
for the probabilistic modeling of images and text (see~\cite{bengio2012representation} for
a review). The work of \citet{srivastava2012multimodal} on DBMs and \citet{sohn2014improved} on MDRNN are  good recent examples. \citet{ngiam2011multimodal}
also proposed deep autoencoder networks for multimodal learning, though this approach was recently shown to be outperformed by DBMs~\cite{srivastava2013discriminative} and MDRNN~\cite{sohn2014improved}.
Although DocNADE shows favorable performance over other topic models, the lack of an efficient deep formulation reduces its ability of modeling multimodal data, especially compared with
the deep neural network based models~\cite{ngiam2011multimodal,srivastava2012multimodal,srivastava2013discriminative}. Thus, the second contribution of this paper is to propose an 
efficient deep version of DocNADE and its supervised variant.
As we'll see, the deep version of our DocNADE model will outperform the DBM
approach of \citet{srivastava2013discriminative}.


\section{Document NADE}
\label{sec: DocNADE intro}


In this section, we describe the original DocNADE model. In \citet{larochelle2012neural},
DocNADE was used to model documents of real words, belonging to some
predefined vocabulary. To model image data, we assume that images have
first been converted into a bag of visual words. A
standard approach is to learn a vocabulary of visual words by
performing $K$-means clustering on SIFT descriptors densely
exacted from all training images. See Section~\ref{experiment:conf} for more details
about this procedure.  From that point on, any image can thus be
represented as a bag of visual words ${\bf v}=[v_1,v_2,\ldots,v_{D_{\bf v}}]$, where each $v_i$ is the index of the
closest $K$-means cluster to the $i^{\rm th}$ SIFT descriptor
extracted from the image and $D_{\bf v}$ is the number of extracted descriptors for image ${\bf v}$.

DocNADE models the joint probability of the visual words $p({\bf v})$ by rewriting it as
\begin{equation}
p\left({\bf v}\right)=\prod_{i=1}^{D_{\bf v}} p\left ( v_i| {\bf v}_{<i}\right ) \label{eqn:prob_chain_rule}
\end{equation}
and modeling instead each conditional $p( v_i| {\bf v}_{<i} )$, where
$\mathbf{v}_{<i}$ is the subvector containing all $v_j$ such
that $j<i$\footnote{ We use a random ordering of the visual words in Equation~\ref{eqn:prob_chain_rule} for each image, and we find it works well in practice. See the discussion in Section~\ref{sec:SupDocNADE} for more details. }.  Notice that Equation~\ref{eqn:prob_chain_rule} is true
for any distribution, based on the probability chain rule. Hence, the
main assumption made by DocNADE is in the form of the conditionals.
Specifically, DocNADE assumes that each conditional can be modeled and
learned by a feedforward neural network.

One possibility would be to model $ p( v_i| \mathbf{v}_{<i}) $ with the following architecture:
\begin{eqnarray}
&&\hspace{-1.5cm}\mathbf{h}_i\left ( \mathbf{v}_{<i} \right ) = {\bf g}\left( \mathbf{c}+\sum_{k<i}\mathbf{W}_{:,v_k} \right ) \label{eqn:docnade_hidden}\\
&&\hspace{-1.5cm} p\left ( v_i=w|\mathbf{v}_{<i} \right ) = \frac{\exp\left ( b_w +\mathbf{V}_{w,:}\mathbf{h}_i\left ( \mathbf{v}_{<i} \right )\right )}{\sum_{w{}'}\exp\left ( b_{w{}'} +\mathbf{V}_{w{}',:}\mathbf{h}_i\left ( \mathbf{v}_{<i} \right )\right )}\label{eqn:docnade_softmax}
\end{eqnarray}
where $g(\cdot)$ is an element-wise non-linear activation function,
$\mathbf{W} \in \mathbb{R}^{H \times Q}$ and $\mathbf{V} \in
\mathbb{R}^{Q \times H}$ are the connection parameter matrices,
$\mathbf{c} \in \mathbb{R}^N$ and $\mathbf{b} \in \mathbb{R}^Q$ are
bias parameter vectors and $H,Q$ are the number of hidden units
(topics) and vocabulary size, respectively.

Computing the distribution $p( v_i=w|\mathbf{v}_{<i} )$ of
Equation~\ref{eqn:docnade_softmax} requires time linear in $Q$. In
practice, this is too expensive, since it must be computed for each of
the $D_{\bf v}$ visual words $v_i$.  To address this issue,
\citet{larochelle2012neural} propose to use a balanced binary tree to
decompose the computation of the conditionals and obtain a complexity
logarithmic in $Q$. This is achieved by randomly assigning all visual
words to a different leaf in a binary tree. Given this tree, the
probability of a word is modeled as the probability of reaching its
associated leaf from the root. \citet{larochelle2012neural} model each left/right transition
probabilities in the binary tree using a set of binary logistic
regressors taking the hidden layer $\mathbf{h}_{i}({\bf v}_{<i})$ as
input. The probability of a given word can then be obtained by
multiplying the probabilities of each left/right choices of the
associated tree path.

Specifically, let $\mathbf{l}\left(v_i\right) $ be the sequence of tree
nodes on the path from the root to the leaf of $v_i$ and let $\pi
\left(v_i\right)$ be the sequence of binary left/right choices 
at the internal nodes along that path. For example, $l\left(v_i\right)_1 $
will always be the root node of the binary tree, and $\pi
\left(v_i\right)_1$ will be $0$ if the word leaf $v_i$ is in the left
subtree or $1$ otherwise. Let $\mathbf{V} \in \mathbb{R}^{T\times
  H} $ now be the matrix containing the logistic regression weights and
$\mathbf{b} \in \mathbb{R}^T$ be a vector containing the 
biases, where $T$ is the number of inner nodes in the binary tree
and $H$ is the number of hidden units. The probability $ p(
  v_i=w|\mathbf{v}_{<i} )$ is now modeled as
\begin{equation}
p( v_i=w|\mathbf{v}_{<i}) = \prod_{k=1}^{|\pi\left(v_i\right)|} p(\pi\left(v_i\right)_k|\mathbf{v}_{<i})~, \label{eqn:docnade_tree}
\end{equation}      
where
\begin{equation}
p(\pi\left(v_i\right)_k=1|\mathbf{v}_{<i})=\textup{ sigm}\left( b_{l\left(v_i\right)_m} +\mathbf{V}_{l\left(v_i\right)_m,:}\mathbf{h}_i\left ( \mathbf{v}_{<i} \right )\right)
\label{eqn:docnade_tree_lr}
\end{equation}
are the internal node logistic regression outputs and $\textup{
  sigm}(x) = 1/(1+\exp(-x))$ is the sigmoid function. By using a
balanced tree, we are guaranteed that computing
Equation~\ref{eqn:docnade_tree} involves only $O(\log_2 Q)$ logistic
regression outputs. One could attempt to optimize the organization of
the words within the tree, but a random assignment of the words to
leaves works well in practice \cite{larochelle2012neural}.

Thus, by combining Equations~\ref{eqn:docnade_hidden},
\ref{eqn:docnade_tree} and \ref{eqn:docnade_tree_lr}, we can compute
the probability $p\left( {\bf v} \right)=\prod_{i=1} p\left ( v_i|{\bf
    v}_{<i}\right ) $ for any document under DocNADE. To train the
parameters $\theta = \lbrace{{\bf W},{\bf V},{\bf b},{\bf c}\rbrace}$
of DocNADE, we simply optimize the average negative log-likelihood of
the training set documents using stochastic gradient descent. 

Equations~\ref{eqn:docnade_tree},\ref{eqn:docnade_tree_lr} indicate that the
conditional probability of each word $v_i$ requires computing the
position dependent hidden layer $\mathbf{h}_i\left( \mathbf{v}_{<i}
\right )$, which extracts a
representation out of the bag of previous visual words
$\mathbf{v}_{<i}$. Since computing $\mathbf{h}_i\left( \mathbf{v}_{<i}
\right )$ is in $O(H D_{\bf v})$ on average, and there are $D_{\bf v}$
hidden layers $\mathbf{h}_i\left( \mathbf{v}_{<i}
\right )$ to compute, then a naive procedure
for computing all hidden layers would be in $O(H D_{\bf v}^2)$.

However, noticing that
\begin{eqnarray}
{\mathbf{h}_{i+1}\left ( \mathbf{v}_{<i+1} \right )} & =& {\bf g}\left( \mathbf{c}+\sum_{k<i+1}\mathbf{W}_{:,v_k} \right )  \\
 &=&  {\bf g}\left( \mathbf{W}_{:,v_i}+\mathbf{c}+\sum_{k<i}\mathbf{W}_{:,v_k} \right )
\end{eqnarray}
and exploiting that fact that the weight matrix $\mathbf{W}$ is the same
across all conditionals, the linear transformation
$\mathbf{c}+\sum_{k<i}\mathbf{W}_{:,v_k}$ can be reused from the computation
of the previous hidden layer $\mathbf{h}_{i}( \mathbf{v}_{<i})$ to compute
$\mathbf{h}_{i+1}( \mathbf{v}_{<i+1})$. With this procedure,
computing all hidden layers $\mathbf{h}_{i}( \mathbf{v}_{<i}
) $ sequentially from $i=1$ to $i=D_{\bf v}$ becomes in $O(H D_{\bf v})$.
 
Finally, since the computation complexity of each of the $O(\log_2 Q)$ logistic
regressions in Equation~\ref{eqn:docnade_tree} is $O(H)$, the total
complexity of computing $ p( v_i=w|\mathbf{v}_{<i} )$ is $O(\log_2(Q) H
D_{\bf v})$. In practice, the length of document $D_{\bf v}$ and the number of hidden
units $H$ tends to be small, while $\log_2(Q)$ will be small even for large
vocabularies. Thus DocNADE can be used and trained efficiently.

Once the model is trained, a latent representation can be extracted from a new
document $\mathbf{v^{\ast}}$ as follows:
\begin{equation}
\mathbf{h}_y\left ( \mathbf{v}^{\ast} \right ) = {\bf g}\left( \mathbf{c}+\sum_{i}^{D_{\bf v}}\mathbf{W}_{:,v^{\ast}_i} \right )~.
\end{equation} 
This representation could be fed to a standard classifier to perform
any supervised computer vision task. The index $y$ is used to
highlight that it is the representation used to predict the class
label $y$ of the image.

\section{SupDocNADE for Multimodal Data}
\label{SupDocNADE intro}
In this section, we describe the approach of this paper, inspired by DocNADE, to
learn jointly from multimodal data. Here, we will concentrate on the single layer version of our model and discuss its deep extension later, in Section~\ref{sec:SupDeepDocNADE}. 

First, we describe a
supervised extension of DocNADE (SupDocNADE), which incorporates the class
label modality into training to learn more discriminative hidden
features for classification. Then we describe how we exploit
the spatial position information of the visual words. Finally,
we describe how to jointly model the text annotation modality with
SupDocNADE. 


\subsection{Supervised DocNADE}
\label{sec:SupDocNADE}
It has been observed that learning image feature representations using
unsupervised topic models such as LDA can perform worse than training
a classifier directly on the visual words themselves, using an
appropriate kernel such as a pyramid
kernel~\cite{lazebnik2006beyond}. One reason is that the unsupervised
topic features are trained to explain as much of the entire
statistical structure of images as possible and might not model
well the particular discriminative structure we are after in our
computer vision task. This issue has been addressed in the literature
by devising supervised variants of LDA, such as Supervised
LDA or sLDA~\cite{blei2007supervised}. DocNADE also being an unsupervised
topic model, we propose here a supervised variant of DocNADE,
SupDocNADE, in an attempt to make the learned image representation
more discriminative for the purpose of image classification.

Specifically, given an image $ {\bf v}=[
v_1,v_2,\ldots,v_{D_{\bf v}}]$ and its class label $y\in \{1,\dots,C\}$,
SupDocNADE models the full joint distribution as
\begin{equation}
p( {\bf v}, y)=p(y|{\bf v})\prod_{i=1}^{D_{\bf v}} p\left ( v_i| {\bf v}_{<i}\right ) ~~. \label{eqn:supdocnade}
\end{equation}
As in DocNADE, each conditional is modeled by a neural network. We use
the same architecture for $p\left ( v_i| {\bf v}_{<i}\right )$ as in regular DocNADE. We now only need
to define the model for $p(y|{\bf v})$.

Since $\mathbf{h}_y\left ( \mathbf{v} \right )$ is the image
representation that we'll use to perform classification, we propose
to model $p\left ( y| {\bf v}\right )$ as a multiclass logistic 
regression output computed from $\mathbf{h}_y\left ( \mathbf{v} \right )$:
\begin{equation}
p\left( y|{\bf v}\right) = {\rm softmax}\left( \mathbf{d} + \mathbf{U}\mathbf{h}_y\left (\mathbf{v} \right) \right)_y\label{eqn:h_class}
\end{equation}
where ${\rm softmax}({\bf a})_i = \exp(a_i) / \sum_{j=1}^C \exp(a_j)$,
 $ \mathbf{d} \in \mathbb{R}^C $ is the bias parameter vector in
the supervised layer and $ \mathbf{U} \in \mathbb{R}^{C \times H} $ is
the connection matrix between hidden layer $\mathbf{h}_y $ and the class label.

Put differently, $p\left ( y| {\bf v}\right )$ is modeled as a regular
multiclass neural network, taking as input the bag of visual words
${\bf v}$. The crucial difference however with a regular neural
network is that some of its parameters (namely the hidden unit
parameters ${\bf W}$ and ${\bf c}$) are also used to model the visual
word conditionals $p\left ( v_i| {\bf v}_{<i}\right )$.

Maximum likelihood training of this model is performed by
minimizing the negative log-likelihood
\begin{equation}
  -\log p\left( {\bf v},y\right) = - \log p\left ( y| {\bf v} \right) +\label{eqn:objectfunc}  \sum_{i=1}^{D_{\bf v}} -\log p( v_i | {\bf v}_{<i})
\end{equation} 
averaged over all training images. This is known as generative
learning~\cite{bouchard2004tradeoff}. The first term is a purely discriminative term,
while the second is unsupervised and can be understood as a
regularizer, that encourages a solution which also explains the
unsupervised statistical structure within the visual words. In
practice, this regularizer can bias the solution too strongly away
from a more discriminative solution that generalizes well. Hence, 
similarly to previous work on hybrid generative/discriminative learning, we propose
instead to weight the importance of the generative term
\begin{equation}
L({\bf v},y;\theta) = - \log p\left ( y| {\bf v} \right) + \lambda \sum_{i=1}^{D_{\bf v}} -\log p( v_i | {\bf v}_{<i}) \label{eqn:objectfunc_hybrid} 
\end{equation} 
where $\lambda$ is treated as a regularization hyper-parameter.

Optimizing the training set average of
Equation~\ref{eqn:objectfunc_hybrid} is performed by stochastic
gradient descent, using backpropagation to compute the parameter
derivatives. As in regular DocNADE, computation of the training objective and its
gradient requires that we define an ordering of the visual
words. Though we could have defined an arbitrary path across the image
to order the words (e.g. from left to right, top to bottom in the
image), we follow~\citet{larochelle2012neural} and randomly permute the words before every
stochastic gradient update. The implication is that the model is
effectively trained to be a good inference model of {\it any} conditional $p(
v_i | {\bf v}_{<i})$, for any ordering of the words in ${\bf v}$. This
again helps fighting against overfitting and better regularizes our
model. One could thus think of SupDocNADE as learning from a sequence of 
\textit{random} fixations performed in a visual scene.

In our experiments, we used the rectified linear function as the
activation function
\begin{equation}
{\bf g}({\bf a}) = \max(0, {\bf a}) = [\max(0,a_1),\dots,\max(0,a_H)]
\end{equation}
which often outperforms other activation functions~\cite{glorot2011deep} and
has been shown to work well for image data~\cite{nair2010rectified}. Since
this is a piece-wise linear function, the (sub-)gradient
with respect to its input, needed by backpropagation to compute the parameter gradients, is simply
\begin{equation}
{\bf 1}_{({\bf g}({\bf a})>0)}= [1_{(g(a_1)>0)},\dots,1_{(g(a_H)>0)}]
\end{equation}
where $1_{P}$ is 1 if $P$ is true and 0 otherwise.

Algorithms~\ref{alg:fprop}~and~\ref{alg:bprop} give pseudocodes
for efficiently computing the joint distribution $p\left({\bf v},y \right)$
and the parameter gradients of Equation~\ref{eqn:objectfunc_hybrid} required
for stochastic gradient descent training.

\begin{algorithm}[t]
\caption{ Computing $p\left({\bf v},y \right)$ using SupDocNADE}
\begin{algorithmic}
\STATE {\bf Input:} bag of words representation ${\bf v}$, target $y$
\STATE {\bf Output:} $p\left({\bf v},y \right)$
\STATE $\mathbf{act}\gets \mathbf{c}$ 
\STATE $p\left(\mathbf{v} \right) \gets 1$ 
\FOR{$i$ from $1$ to $D_{\bf v}$}
  \STATE $\mathbf{h}_i \gets$ ${\bf g}\left( \mathbf{act}\right)$
  \STATE $p\left(v_i|\mathbf{v}_{<i}\right)=1$ 
  \FOR{$m$ from 1 to $|\pi \left(v_i\right)|$}
    \STATE  $p\left(v_i|\mathbf{v}_{<i}\right) \gets  p\left(v_i|\mathbf{v}_{<i}\right) p\left(\pi\left(v_i\right)_m|\mathbf{v}_{<i}\right)$
  \ENDFOR
  \STATE $p\left(\mathbf{v} \right) \gets p\left(\mathbf{v} \right)p\left(v_i|\mathbf{v}_{<i}\right)$ 
  \STATE $\mathbf{act}\gets \mathbf{act} + \mathbf{W}_{:,v_i}$ 
\ENDFOR
\STATE $\mathbf{h}^{c}\left (\mathbf{v} \right ) \gets \max(0,\mathbf{act}) $
\STATE $p\left( y|\mathbf{v}\right) \gets \textup{softmax} \left( \mathbf{d} + \mathbf{U}\mathbf{h}^{c}\left (\mathbf{v} \right ) )\right)_{|y}$
\STATE $p\left({\bf v},y \right) \gets p\left(\mathbf{v} \right)p\left( y|\mathbf{v}\right) $
\end{algorithmic}
\label{alg:fprop}
\end{algorithm}

\begin{algorithm}[t]
\caption{ Computing SupDocNADE training gradients}
\begin{algorithmic}
\STATE {\bf Input:} training vector ${\bf v}$, target $y$,\\
\hspace{10mm} unsupervised learning weight $\lambda$
\STATE {\bf Output:} gradients of Equation~\ref{eqn:objectfunc_hybrid} w.r.t. parameters
\STATE $f\left(\mathbf{v}\right) \gets \textup{softmax} \left( \mathbf{d} + \mathbf{U}\mathbf{h}^{c}\left (\mathbf{v} \right ) )\right)$
\STATE $\delta \mathbf{d} \gets \left(f \left(\mathbf{v}\right)-1_y\right)$
\STATE $\delta \mathbf{act} \gets (\mathbf{U}^\intercal \delta \mathbf{d})  \circ 1_{{\bf h}_y > 0}$
\STATE $\delta \mathbf{U} \gets \delta \mathbf{d}~{\mathbf{h}^{c^\intercal}}$
\STATE $\delta \mathbf{c} \gets 0$, $\delta \mathbf{b} \gets 0$, $\delta \mathbf{V} \gets 0$, $\delta \mathbf{W} \gets 0$
\FOR{$i$ from $D_{\bf v}$ to $1$}
   \STATE $\delta\mathbf{h}_i \gets 0$
   \FOR {$m$ from $1$ to $|\pi\left(v_i\right)|$}
   \STATE $ \delta t \gets \lambda \left(p\left(\pi\left(v_i\right)_m|\mathbf{v}_{<i}\right)-\pi\left(v_i\right)_m\right)$
   \STATE $\delta b_{l\left(v_i\right)_m} \gets \delta b_{l\left(v_i\right)_m}+\delta t$
   \STATE $\delta \mathbf{V}_{l\left(v_i\right)_m,:} \gets \delta \mathbf{V}_{l\left(v_i\right)_m,:}+ \delta t~\mathbf{h}_i^\intercal$
   \STATE $\delta \mathbf{h}_i \gets \delta \mathbf{h}_i + \delta t~\mathbf{V}_{l\left(v_i\right)_m,:}^\intercal$
   \ENDFOR
\STATE $\delta \mathbf{act} \gets \delta \mathbf{act}+ \delta \mathbf{h}_i\circ 1_{{\bf h}_i > 0}$
\STATE $\delta \mathbf{c} \gets \delta \mathbf{c} + \delta \mathbf{h}_i\circ 1_{{\bf h}_i > 0}$
\STATE $\delta \mathbf{W}_{:,v_i} \gets \delta \mathbf{W}_{:,v_i} + \delta \mathbf{act}$
\ENDFOR

\end{algorithmic}
\label{alg:bprop}
\end{algorithm}

\subsection{Dealing with Multiple Regions}
\label{sec:multiple regions}
Spatial information plays an important role for understanding an
image. For example, the sky will often appear on the top part
of the image, while a car will most often appear at the
bottom. A lot of previous work has exploited this intuition
successfully. For example, in the seminal work on spatial
pyramids~\cite{lazebnik2006beyond}, it is shown that extracting
different visual word histograms over distinct regions instead of a
single image-wide histogram can yield substantial gains in
performance.

We follow a similar approach, whereby we model both the presence of the
visual words and the identity of the region they appear in. 
Specifically, let's assume the image is divided into several distinct regions
${\cal R} = \lbrace{ R_1,R_2, \ldots, R_M \rbrace }$, where $M$ is the number of
regions. The image can now be represented as
\begin{eqnarray}
{\bf v}^{\cal R}& =& [ v^{\cal R}_1,v^{\cal R}_2, \ldots, v^{\cal R}_D ] \\
& =& [ \left( v_1,r_1 \right),\left( v_2, r_2 \right),\ldots,\left(v_{D_{\bf v}},r_{D_{\bf v}} \right) ]\nonumber
\end{eqnarray} 
where $r_i \in {\cal R}$ is the region
from which the visual word $v_i$ was extracted. To model the joint
distribution over these visual words, we decompose it as $p({\bf v}^{\cal R}) =
\prod_i p((v_i,r_i) | {\bf v}^{\cal R}_{<i})$ and treat each $Q\times M$
possible visual word/region pair as a distinct word. One implication
of this is that the binary tree of visual words must be larger so as
to have a leaf for each possible visual word/region pair. Fortunately,
since computations grow logarithmically with the size of the tree,
this is not a problem and we can still deal with a large
number of regions.

\subsection{Dealing with Annotations}
\label{sec: anno}
So far, we've described how to model the visual word and class label modalities.
In this section, we now describe how we also model the annotation word modality with
SupDocNADE. 

Specifically, let ${\cal A}$ be the predefined vocabulary of all
annotation words, we will note the annotation of a given image as $
{\bf a}= [ a_1,a_2, \ldots ,a_L ] $ where $a_i \in
{\cal A}$, with $L$ being the number of words in the annotation. Thus,
the image with its annotation can be represented as a mixed bag of visual
and annotation words:
\begin{eqnarray}
{\bf v}^{\cal A} & = & [ v_1^{\cal A},\ldots,v_{D_{\bf v}}^{\cal A} , v_{D_{\bf v}+1}^{\cal A}, \ldots, , v_{D_{\bf v}+L}^{\cal A}] \\
& = & [ v^{\cal R}_1,\ldots,v^{\cal R}_{D_{\bf v}} , a_1, \ldots ,a_L ]~~. \nonumber
\end{eqnarray}
To embed the annotation words into the SupDocNADE framework, we treat
each annotation word the same way we deal with visual
words. Specifically, we use a joint indexing of all visual and
annotation words and use a larger binary word tree so as to augment it with
leaves for the annotation words. By training SupDocNADE on this
joint image/annotation representation ${\bf v}^{\cal A}$, 
it can learn the relationship between the labels, the spatially-embedded 
visual words and the annotation words.

At test time, the annotation words are not given and we wish to
predict them. To achieve this, we compute the document representation
${\bf h}_y({\bf v}^{\cal R})$ based only on the visual words and
compute for each possible annotation word $a \in {\cal A}$ the
probability that it would be the next observed word $p(v_i^{\cal A} = a
|{\bf v}^{\cal A} = {\bf v}^{\cal R})$, based on the tree decomposition as in 
Equation~\ref{eqn:docnade_tree}. In other words, we only compute the probability
of paths that reach a leaf corresponding to an annotation word (not a visual word).
We then rank the annotation words in ${\cal A}$ in decreasing order of their probability
and select the top 5 words as our predicted annotation. 
\section{Deep Extension of SupDocNADE}
\label{sec:SupDeepDocNADE}
Although SupDocNADE has achieved better performance than the other topic models in our
previous work  \cite{zhengtopic}, the lack of an efficient deep formulation of SupDocNADE reduces its capability of modeling 
multimodal data, especially compared with other models based on deep neural network~\cite{srivastava2013discriminative, srivastava2012multimodal}. 

Recently, \citet{uria2013deep} proposed
an efficient extension of the original NADE model~\cite{larochelle2011neural} for binary vector observations, from which DocNADE was derived. We take inspiration from \citet{uria2013deep} and propose SupDeepDocNADE, i.e.\  
a supervised deep autoregressive neural topic model for multimodal data modeling. 

In this section, we introduce the deep extension of DocNADE (DeepDocNADE) and then describe how to incorporate supervised information into its training. 
We  also discuss how to deal with the inbalance between the number of visual words and 
annotation words, in order to obtain good performances. Before we start the discussion, we note that the notation $\bf v$, which denotes the words of an image,
includes both visual words and annotation words of an image in the following section, as is discussed in Section~\ref{sec: anno}

\subsection{DocNADE revisited}
\label{sec: docnade_revisit}

We first revisit the training procedure for DocNADE. We will concentrate on the unsupervised version of DocNADE for now and discuss the supervised case later. 

In Section~\ref{sec:SupDocNADE} we mentioned that words are randomly permuted before every stochastic gradient update, to
make DocNADE be a good inference model for any ordering of the words. 
As \citet{uria2013deep} notice, we can think of the use of many orderings as the instantiation of many different DocNADE models, one for each distinct ordering. From that point of view, by training a single set of parameters (connection matrices and biases) on all these orderings, we are effectively
employing a parameter sharing strategy across these models and the training process can be interpreted as training a factorial number 
of DocNADE models simultaneously. 

We will now make the notion of ordering more explicit in our notation. Following \citet{uria2013deep}, we now denote $p\left({\bf v}|{\bf \theta}, o\right)$ as the joint distribution of the DocNADE model over the words of an image  given the parameters $\mathbf{\theta}$ and 
ordering $o$. We will also note $p\left(v_{o_{d}}|{\bf v}_{o_{<d}}, {\bf \theta}, o\right)$ as the conditional distribution described in Equation~\ref{eqn:docnade_softmax} or 
\ref{eqn:docnade_tree}, where ${\bf v}_{o_{<d}}$ is the subvector of the previous $d-1$ words extracted from an ordered word vector ${\bf v}_{o}$, and $v_{o_{d}}$
is the $d^{\rm th}$ word of ${\bf v}_{o}$. Notice that the ordering $o$ is now treated explicitly as a random variable. 

Thus, training DocNADE on stochastically sampled orderings
corresponds, in expectation, to minimize the negative log-likelihood $-\log p\left({\bf v}|{\bf \theta}, o\right) $ across {\it all possible orderings}, for each training example ${\bf v}$:
\begin{equation}
L\left({\bf v};{\bf \theta}\right) = \underset{o\in \mathcal{O}}{\mathbb{E}}-\log p\left({\bf v}|{\bf \theta}, o\right) \label{eqn: DocNADE_ord}
\end{equation}
where $\mathcal{O}$ is the set of all orderings.

Applying DocNADE's autoregressive expression for the conditionals in Equation~\ref{eqn:prob_chain_rule}, Equation~\ref{eqn: DocNADE_ord} can be rewritten as:
\begin{equation}
L\left({\bf v};{\bf \theta}\right) = \underset{o\in \mathcal{O}}{\mathbb{E}}\sum_{d} -\log p\left(v_{o_{d}}|{\bf v}_{o_{<d}}, {\bf \theta}, o\right) \label{eqn: DocNADE_autoreg}
\end{equation}

By moving the expectation over orderings, $\underset{o\in \mathcal{O}}{\mathbb{E}}$, inside the summation over the conditionals, the expectation can be split into three parts\footnote{ The split is done in a modality-agnostic way, i.e. the visual words and annotations words are mixed together and are treated equally when training the model.}:
one over $o_{<d}$, standing for the first $d-1$ indices in the ordering $o$; one over $o_d$, which is the $d^{\rm th}$ index of the ordering $o$; and one over $o_{>d}$, standing for the remaining indices of 
the ordering. 

Hence, the loss function can be rewritten as:
\begin{equation}
L\left({\bf v};{\bf \theta}\right) = \sum_{d}\underset{o_{<d}}{\mathbb{E}}
\underset{o_{d}}{\mathbb{E}}\underset{o_{>d}}{\mathbb{E}} -\log p\left(v_{o_{d}}|{\bf v}_{o_{<d}}, {\bf \theta}, o_{<d}, o_d, o_{>d}\right) \label{eqn: DocNADE_split}
\end{equation} 

Noting that the value of each conditional does not depend on $o_{>d}$, Equation~\ref{eqn: DocNADE_split} can then be simplified as:
\begin{equation}
L\left({\bf v};{\bf \theta}\right) = \sum_{d}\underset{o_{<d}}{\mathbb{E}}
\underset{o_{d}}{\mathbb{E}} -\log p\left(v_{o_{d}}|{\bf v}_{o_{<d}}, {\bf \theta}, o_{<d}, o_d\right)~. \label{eqn: DocNADE_split_simplified}
\end{equation} 

In practice, Equation~\ref{eqn: DocNADE_split_simplified} still sums over a number of terms of too large to be performed exhaustively. For training, we thus use a stochastic estimation and replace the expectations/sums over $d$ and $o_{<d}$ with samples. On the other hand, the innermost expectation over $o_{d}$ can
be obtained cheaply. Indeed, for a given value of $d$ and $o_{<d}$, all terms $p\left(v_{o_{d}}|{\bf v}_{o_{<d}}, {\bf \theta}, o_{<d}, o_d\right)$ require the computation of the same hidden layer representation ${\bf h}_d \left({\bf v}_{o_{<d}}\right)$ from the subvector ${\bf v}_{o_{<d}}$. Therefore, 
$L\left({\bf v},{\bf \theta}\right)$ can be estimated by:
\begin{equation}
\hat{L}\left({\bf v},{\bf \theta}\right) = \frac{D_{\bf v}}{D_{\bf v}-d+1}\sum_{o_d}-\log p\left(v_{o_{d}}|{\bf v}_{o_{<d}}, {\bf \theta}, o_{<d}, o_d\right)\label{eqn:DocNADE_est}
\end{equation}
where $D_{\bf v}$ is the number of words (including both visual and annotation words) in $\mathbf{v}$. 
In words, Equation~\ref{eqn:DocNADE_est} measures the ability of the model to predict, from a fixed and random context of $d-1$ words ${\bf v}_{o<d}$, any of the remaining words in the image/annotation.

From this, training of DocNADE can be performed by stochastic gradient descent. For a given training example ${\bf v}$, a training update is performed as follows\footnote{In experiments, both visual words and annotation words are represented in Bag of Words (BoW) fashion. As is shown in Section~\ref{sec: DeepDocNADE}, the  training processing actually equals to generating a word vector $\mathbf{v}$ from BoW, shuffling the word vector $\bf{v}$ and splitting it, and then regenerating the histogram $\mathbf{x}\left({\bf v}_{o_{<d}}\right)$ and $\mathbf{x}\left({\bf v}_{o_{\geq d}}\right)$, which is inefficient for processing samples in a mini-batch fashion. Hence, in practice, we split the original histogram $\mathbf{x}\left({\bf v}\right)$ directly by uniformly sampling how many are put in the left of the split (the others are put on the right of the split) for each individual word. This is not equivalent to the one mentioned in this paper, but it works well in practice. }: 
\begin{itemize}
\item[$1)$.] Shuffle ${\bf v}$ to specify an ordering $o$; 
\item[$2)$.] Sample $d$ uniformly from $\left[0, D_{\bf v}\right]$, which separates ${\bf v}$ into two parts: $\mathbf{v}_{o_{<d}}$ as inputs and $\mathbf{v}_{o_{\geq d}}$ as outputs;
\item[$3)$.] Compute each of the conditionals in Equation~\ref{eqn:DocNADE_est}, where $o_d \in \mathbf{v}_{o_{\geq d}} $ ;
\item[$4)$.] Compute and sum the gradients for each of the conditionals in Equation~\ref{eqn:DocNADE_est}, and rescale by $\frac{D_{\bf v}}{D_{\bf v}-d+1}$.
\end{itemize}

It should be noticed that, since the number of words
in an image/annotation pair can vary across examples, the
value of $D_{\bf v}$ will vary between updates, unlike
in \citet{uria2013deep} will models binary vectors of fixed
size.

We can contrast this procedure from the one described in Section~\ref{sec:SupDocNADE}, which prescribed a stochastic estimation with respect to the possible orderings of the words and an exhaustive sum in predicting all the words in the sequence. Here, we have the opposite: it is stochastic by predicting a subset of the words but is (partially) exhaustive by implicitly summing the gradient contributions over several orderings sharing the same permutation up to position $d$.


\subsection{Deep Document NADE}
\label{sec: DeepDocNADE}

As shown in Section~\ref{sec: docnade_revisit}, training of DocNADE can be performed by randomly splitting the words $\mathbf{v}$ into two parts,
$\mathbf{v}_{o_{<d}}$ and $\mathbf{v}_{o_{\geq d}}$, and applying stochastic gradient descent on the loss function of Equation~\ref{eqn:DocNADE_est}. Thus, the training procedure
 now corresponds to a neural network, with $\mathbf{v}_{o_{<d}}$ being the input and $\mathbf{v}_{o_{\geq d}}$ as the output's target. The advantage of this approach is that DocNADE can more easily be extended to a deep version this way, which we will refer to as DeepDocNADE.

Indeed, as mentioned in the previous section, all conditionals $p\left(v_{o_{d}}|\mathbf{v}_{o_{<d}}, \theta,o_{<d}, o_{d}\right)$ in the summation of Equation~\ref{eqn:DocNADE_est}
require the computation of a single hidden layer representation:
\begin{eqnarray}
\mathbf{h}_d^{\left(1\right)}\left({\bf v}_{o<d}\right) &= {\bf g}\left( \mathbf{c}^{\left(1\right)}+\sum_{k<d}\mathbf{W}^{\left(1\right)}_{:,v_{o_k}} \right )\\
    &= {\bf g}\left({\bf c}^{\left(1\right)} + {\bf W}^{\left(1\right)}\mathbf{x}\left({\bf v}_{o_{<d}}\right) \right) \label{eqn: deepdocnade_h1}
\end{eqnarray}
where $\mathbf{x}\left({\bf v}_{o_{<d}}\right)$ is the histogram vector representation
of the word sequence ${\bf v}_{o_{<d}}$ and where the exponent $(1)$ is used to index the first hidden layer and its parameters.

So, unlike in the original training procedure for DocNADE, a training update now requires the computation of a single hidden layer, instead of $D_{\bf v}$ hidden layers. This way, adding more hidden layers only has an additive, instead of multiplicative, effect on the complexity of each training update. 
Hidden layers are added as in regular deep feedforward neural networks, as follows:
\begin{equation}
\mathbf{h}^{\left(n\right)} = {\bf g}\left({\bf c}^{\left(n\right)} + {\bf W}^{\left(n\right)}\mathbf{h}^{\left(n-1\right)} \right) \label{eqn: deepdocnade_h}
\end{equation}
where ${\bf W}^{\left(n\right)}$ and ${\bf c}^{\left(n\right)}$ are the connection matrix and bias for hidden layer $\mathbf{h}^{\left(n\right)}$, $n=1,\ldots, N$,  
where $N$ is the number of hidden layers.

To compute the conditional $p\left(v_{o_{d}}|{\bf v}_{o_{<d}}, {\bf \theta}, o_{<d}, o_d\right)$ in Equation~\ref{eqn:DocNADE_est} after obtaining
the hidden representation $\mathbf{h}^{\left(N\right)}$, the binary tree introduced in Section~\ref{sec: DocNADE intro} could be used for an
efficient implementation. However, in cases where the histogram $\mathbf{x}\left({\bf v}_{o_{\geq d}}\right)$ of future words is not sparse, the binary tree output model might not be the most efficient approach.
For example, suppose $\mathbf{x}\left({\bf v}_{o_{\geq d}}\right)$ is full (has no zero entries) and the vocabulary size is
$Q$, the computation of Equation~\ref{eqn:DocNADE_est} via the binary tree is in $O\left(Q\log_2 Q\right)$,  since it has to compute $O\left(\log Q\right)$ logistic regressions
for each of the $Q$ words in $\mathbf{x}\left({\bf v}_{o_{\geq d}}\right)$. In this specific scenario however, going back to a softmax model of the conditionals is preferable. Indeed, since all conditionals in Equation~\ref{eqn:DocNADE_est} share the same 
hidden representation $\mathbf{h}^{\left(N\right)}$ and thus the normalization term in
the softmax is the same for all future words, it is only in $O\left(Q\right)$. Another advantage of the softmax over the binary tree is that the softmax is more amenable to an efficient implementation on the GPU, which will also speed up the training process. 

In the end, for the experiments with the deep extension of DocNADE of this paper, we opted for the softmax model as we've found it to be more efficient. 
We emphasize however that the binary tree is still the most efficient option
for the loss function of Equation~\ref{eqn:objectfunc_hybrid} or when the histogram of future words $\mathbf{x}\left({\bf v}_{o_{\geq d}}\right)$ is sparse. 

\subsection{Supervised Deep Document NADE}
Deep Document NADE can also be extended to a supervised variant, which is referred to as SupDeepDocNADE, following the formulation in Section~\ref{sec:SupDocNADE}. 

Specifically, to add the supervised information into DeepDocNADE, the negative log-likelihood function in Equation~\ref{eqn: DocNADE_ord} could be extended as follows:
\begin{eqnarray}
L\left({\bf v},y;{\bf \theta}\right) &=& \underset{o\in \mathcal{O}}{\mathbb{E}}-\log p\left({\bf v}, y|{\bf \theta}, o\right)  \\
 &=&  \underset{o\in \mathcal{O}}{\mathbb{E}}-\log p\left(y| {\bf v}, {\bf \theta}\right)-\log p\left({\bf v}|{\bf \theta}, o\right) \label{eqn: supdocnade_ord}
\end{eqnarray}

Since $p\left(y| {\bf v}, {\bf \theta}\right)$ is independent of $o$, Equation~\ref{eqn: supdocnade_ord} can be rewritten as:
\begin{equation}
L\left({\bf v},y;{\bf \theta}\right)=-\log p\left(y| {\bf v}, {\bf \theta}\right)-\underset{o\in \mathcal{O}}{\mathbb{E}}\log p\left({\bf v}|{\bf \theta}, o\right) \label{eqn: supdocnade_split}
\end{equation} 

Then $L\left({\bf v},y;{\bf \theta}\right)$ can be approximated by sampling ${\bf v}$, $d$ and $o_{<d}$ as follows:
\begin{eqnarray}
\hat{L}\left({\bf v},y;{\bf \theta}\right) &=& -\log p\left(y| {\bf v}, {\bf \theta}\right)\label{eqn:SupDocNADE_est}\\ \nonumber
&-&\frac{D_{\bf v}}{D_{\bf v}-d+1}\sum_{o_d}\log p\left(v_{o_{d}}|{\bf v}_{o_{<d}}, {\bf \theta}, o_{<d}, o_d\right)
\end{eqnarray}

Similar to Equation~\ref{eqn:objectfunc_hybrid}, the first term in Equation~\ref{eqn:SupDocNADE_est} is supervised, while the second term is unsupervised and can be interpreted as a regularizer. Thus, we can also weight the importance of the unsupervised part by a hyperparameter $\lambda$ 
and obtain a hybrid cost function:
\begin{eqnarray}
\hat{L}\left({\bf v},y;{\bf \theta}\right) &=& -\log p\left(y| {\bf v}, {\bf \theta}\right)\label{eqn:SupdeepDocNADE_hybrid}\\ \nonumber
&-&\lambda\frac{D_{\bf v}}{D_{\bf v}-d+1}\sum_{o_d}\log p\left(v_{o_{d}}|{\bf v}_{o_{<d}}, {\bf \theta}, o_{<d}, o_d\right)
\end{eqnarray}
Equation~\ref{eqn:SupdeepDocNADE_hybrid} can then be used as the per-example loss and optimized over the training set using stochastic gradient descent.

\subsection{Weighting the Annotation Words}
\label{sec:weight_anno}
As mentioned in Section~\ref{sec: anno}, the annotation words can be embedded into the framework of SupDocNADE by treating them the same way we deal with
visual words. In practice, however, the number of visual words could be much larger than that of the annotation words. For example, in the MIR Flickr data set, 
with the experimental setup of \citet{srivastava2012multimodal}, the average number of visual words for an image is about $69~011$, which is much larger than the average number of annotation words for an image ($5.15$). The imbalance of visual words and annotation words might cause some problems. For example, 
the contribution to the hidden representation from the annotation words is so small that it might be ignored compared with the contribution from the huge mount of
visual words, and the gradients coming from the annotation words might also be too small to have any meaningful effect for increasing the conditionals 
probability of the annotation words. 

To deal with this problem, we propose to weight the annotation words in
the histogram $\mathbf{x}\left({\bf v}_{o_{<d}}\right)$ and $\mathbf{x}\left({\bf v}_{o_{\geq d}}\right)$. More specifically, let 
$\omega\left(\rho\right) \in \mathbb{R}^{Q}$ be a vector containing $Q$ components, where $Q$ is the vocabulary size (including both visual and annotation words),
each component corresponding to a word (either visual or annotation). The components corresponding to the visual words is set to $1$ and the components
corresponding to the annotation word is set to $\rho$. Then the new histogram of $\mathbf{\tilde{x}}\left({\bf v}_{o_{<d}}\right)$ and $\mathbf{\tilde{x}}\left({\bf v}_{o_{\geq d}}\right)$ is computed as
\begin{eqnarray}
\mathbf{\tilde{x}}\left({\bf v}_{o_{<d}}\right) &=& \mathbf{x}\left({\bf v}_{o_{<d}}\right) \odot \omega\left(\rho\right)\\
\mathbf{\tilde{x}}\left({\bf v}_{o_{\geq d}}\right) &=& \mathbf{x}\left({\bf v}_{o_{\geq d}}\right) \odot \omega\left(\rho\right)
\end{eqnarray}
where $\odot$ is element-wise multiplication. 

Moreover, the hybrid cost function of Equation~\ref{eqn:SupdeepDocNADE_hybrid} is rewritten as:
\begin{eqnarray}
&&\hat{L}\left({\bf v},y;{\bf \theta}\right) = -\log p\left(y| {\bf v}, {\bf \theta}\right)\label{eqn:SupdeepDocNADE_hybri_weight}\\ \nonumber
&&~~~~-\frac{\lambda D_{\bf v}}{D_{\bf v}-d+1}\sum_{o_d}\Phi_{o_{d}}\left(\rho\right)\log\tilde{p}\left(v_{o_{d}}|{\bf v}_{o_{<d}}, {\bf \theta}, o_{<d}, o_d\right)
\end{eqnarray}
where $\tilde{p}\left(v_{o_{d}}|{\bf v}_{o_{<d}}, {\bf \theta}, o_{<d}, o_d\right)$ is a conditional probability obtained by replacing $\mathbf{x}\left({\bf v}_{o_{<d}}\right)$
with $\mathbf{\tilde{x}}\left({\bf v}_{o_{<d}}\right)$ in Equation~\ref{eqn: deepdocnade_h1}, and $\Phi_{o_{d}}\left(\rho\right)$ is a function that assigns
weight $\rho$ if $o_{d}$ is an annotation word, and $1$ otherwise.

By weighting annotation words in the histogram, the model will pay more attention to the annotation words, reducing the problem caused by the imbalance between visual and annotation words. In practice, the weight $\rho$ is a hyper-parameter and can be selected by cross-validation. As we'll see in Section~\ref{sec:SupDeepDocNADE annoweight}, weighting annotation words more heavily can significantly improve the performance. 
 
\subsection{Exploiting Global Image Features}
\label{sec:global_features}
Besides the spatial information and annotation which are embedded into the framework of DocNADE in Section~\ref{sec:multiple regions} and Section~\ref{sec: anno},
 bottom-up global features, such as Gist~\cite{oliva2001modeling} and MPEG-7 descriptors~\cite{manjunath2001color}, can also play an important role in
multimodal data modeling \citep{srivastava2012multimodal}. Global features can, among other things, complement the local information extracted from patch-based visual words. In this section, we describe how to embed such features into the framework of 
our model.

Specifically, let $\mathbf{f}\in \mathbb{R}^{N_f}$ be the global feature vector extracted from an image, where $N_f$ is the length of the global feature vector. One
possibility for embedding $\mathbf{f}$ into the model could be to condition the hidden representation on the global feature $\mathbf{f}$ as follows:
\begin{equation}
\mathbf{h}^{\left(1\right)} = {\bf g}\left({\bf c}^{\left(1\right)} + {\bf W}^{\left(1\right)}\mathbf{x}\left({\bf v}_{o_{<d}}\right)
+\mathbf{P}\mathbf{f} \right) \label{eqn:global_feature}
\end{equation}
where $\mathbf{P}$ is a connection matrix specific to the global features.
This can be understood as a hidden layer whose hidden unit biases are conditioned on the image's global features vector $\mathbf{f}$. 
Thus, the whole model is conditioned not only on previous words but also on the global features $\mathbf{f}$.




\section{Experiments and Results}
\label{experiment}
In this section, we compare the performance of our model over the other models for multimodal data modeling. Specifically, 
we first test the ability of the single hidden layer SupDocNADE to learn from multimodal data on two real-world data sets which are widely used in the research on other topic models. Then we test the performance of SupDeepDocNADE on the largescale multimedia informaton retrieval (MIR) Flickr data set and show that SupDeepDocNADE achieves
state-of-the-art performance.
The code to
download the data sets and for SupDocNADE and SupDeepDocNADE is available at
\url{https://sites.google.com/site/zhengyin1126/home/supdeepdocnade}.

\subsection{Experiments for SupDocNADE}
\label{sec:expsupdocnade}
To test the ability of the single hidden layer SupDocNADE to learn from multimodal data, 
we measured its performance under simultaneous image classification and annotation tasks.
We tested our model on 2 real-world data sets: a subset
of the LabelMe data set~\cite{russell2008labelme} and the UIUC-Sports data set~\cite{li2007and}. 
LabelMe and UIUC-Sports come
with annotations and are popular classification and annotation
benchmarks. We performed extensive quantitative comparisons of SupDocNADE with
the original DocNADE model and supervised LDA
(sLDA)\footnote{We mention that \cite{wang2009simultaneous} has shown that sLDA performs better than
Corr-LDA\cite{blei2003modeling}. Moreover, \cite{jia2011learning} found
that Multimodal LDA~\cite{putthividhy2010topic} did not improve on the performance
of Corr-LDA. Finally, sLDA distinguishes itself from the other models
in the fact that it also supports the class label modality and has code available
online. Hence, we compare directly with sLDA only.}~\cite{blei2007supervised,wang2009simultaneous}.
We also provide some comparisons with MMLDA~\cite{wang2011max} and a Spatial Pyramid Matching (SPM) approach~\cite{lazebnik2006beyond}.

\subsubsection{Data sets Description}

 
Following \citet{wang2009simultaneous}, we constructed our LabelMe
data set using the online tool to obtain images of size $256 \times 256$ pixels from the
following 8 classes: {\it highway}, {\it inside city}, {\it coast}, {\it forest},
{\it tall building}, {\it street}, {\it open country} and {\it mountain}. For each
class, 200 images were randomly selected and split evenly in the training
and test sets, yielding a total of 1600 images. 

The UIUC-Sports data set contains 1792 images, classified into
8 classes: {\it badminton} (313 images), {\it bocce} (137 images), {\it croquet}
(330 images), {\it polo} (183 images), {\it rockclimbing} (194 images), {\it rowing}
(255 images), {\it sailing} (190 images), {\it snowboarding} (190
images). Following previous work, the maximum side of each image was
resized to 400 pixels, while maintaining the aspect ratio. We
randomly split the images of each class evenly into training and test sets. 
For both LabelMe and UIUC-Sports
data sets, we removed the annotation words occurring less than 3 times, as
in \citet{wang2009simultaneous}.

\subsubsection{Experimental Setup for SupDocNADE}
\label{experiment:conf}

Following \citet{wang2009simultaneous}, 128
dimensional, densely extracted SIFT features were used to extract the
visual words. The step and patch size of the dense SIFT extraction was
set to 8 and 16, respectively. The dense SIFT features from the
training set were quantized into 240 clusters, to construct our visual
word vocabulary, using $K$-means.  We divided each image into
a $2 \times 2 $ grid to extract the spatial position information, as described in
Section~\ref{sec:multiple regions}.  This produced
$2\times 2\times 240=960$ different visual word/region pairs.

We use classification accuracy to evaluate the performance of
image classification and the average F\textup{-measure} of the top
5 predicted annotations to evaluate the annotation performance, as
in previous work. The F\textup{-measure} of an image is defined as
\begin{equation}
F\textup{-measure} = \frac{2\times \textup{Precision}\times \textup{Recall}}{ \textup{Precision}+\textup{Recall}}
\end{equation}
where recall is the percentage of correctly predicted annotations out
of all ground-truth annotations for an image, while the precision is
the percentage of correctly predicted annotations out of all predicted
annotations\footnote{When there are repeated words in the ground-truth
  annotations, the repeated terms were removed to calculate the
  F\textup{-measure}.}. We used 5 random train/test splits to estimate
the average accuracy and F\textup{-measure}.

Image classification with SupDocNADE is performed by feeding the learned
document representations to a RBF kernel SVM. In our experiments, all
hyper-parameters (learning rate, unsupervised learning weight
$\lambda$ in SupDocNADE, $C$ and $\gamma$ in RBF kernel SVM), were
chosen by cross validation. We emphasize that, again from following \citet{wang2009simultaneous}, the annotation words are
not available at test time and all methods predict an image's class
based solely on its bag of visual words.

\subsubsection{Quantitative Comparison}

In this section, we describe our quantitative comparison between
SupDocNADE, DocNADE and sLDA. We used the implementation of sLDA
available at \url{http://www.cs.cmu.edu/~chongw/slda/} in our
comparison, to which we fed the same visual (with spatial regions) and
annotation words as for DocNADE and SupDocNADE.


%


\begin{figure*}[t]
  \includegraphics[width=0.24\linewidth]{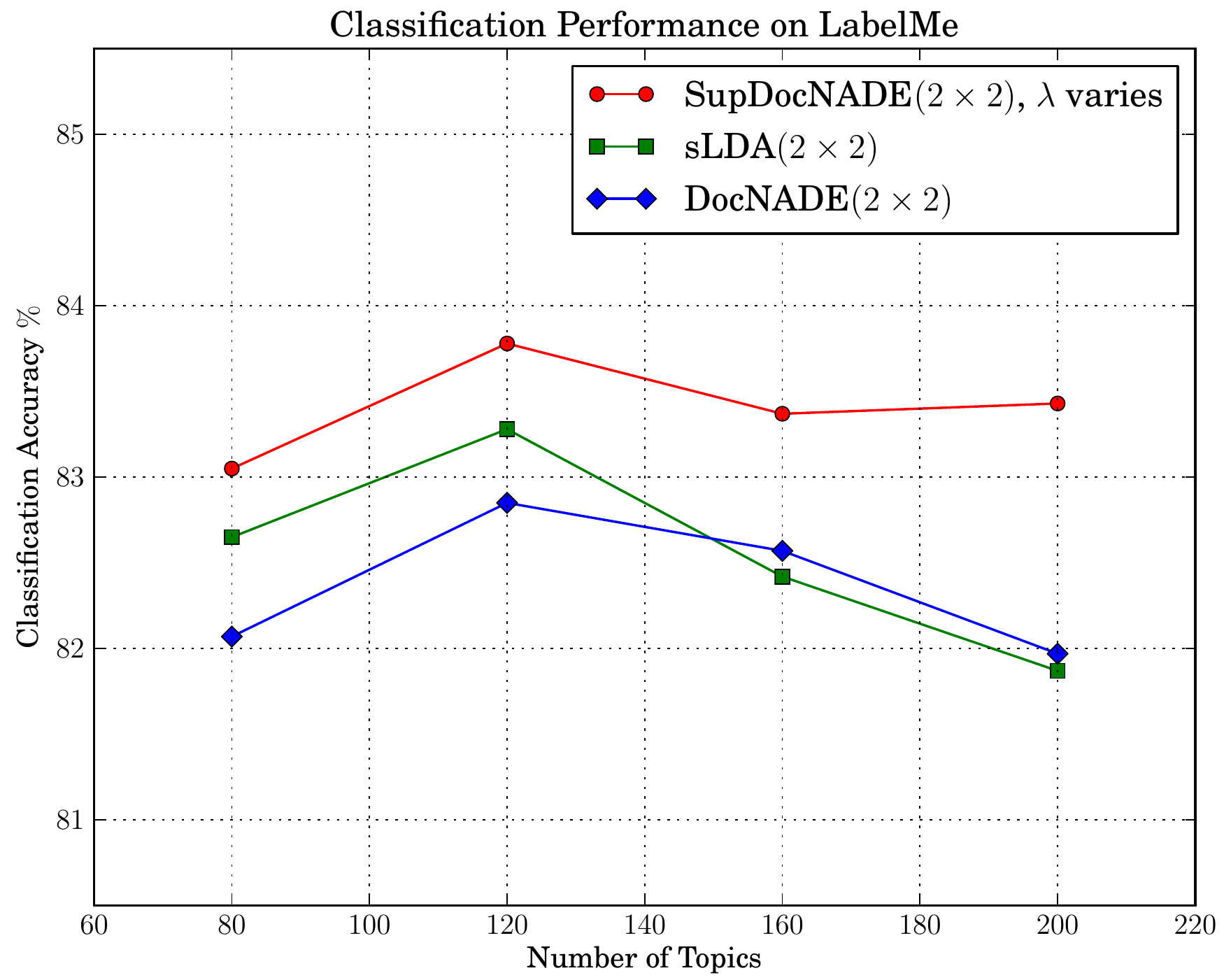}
  \includegraphics[width=0.24\linewidth]{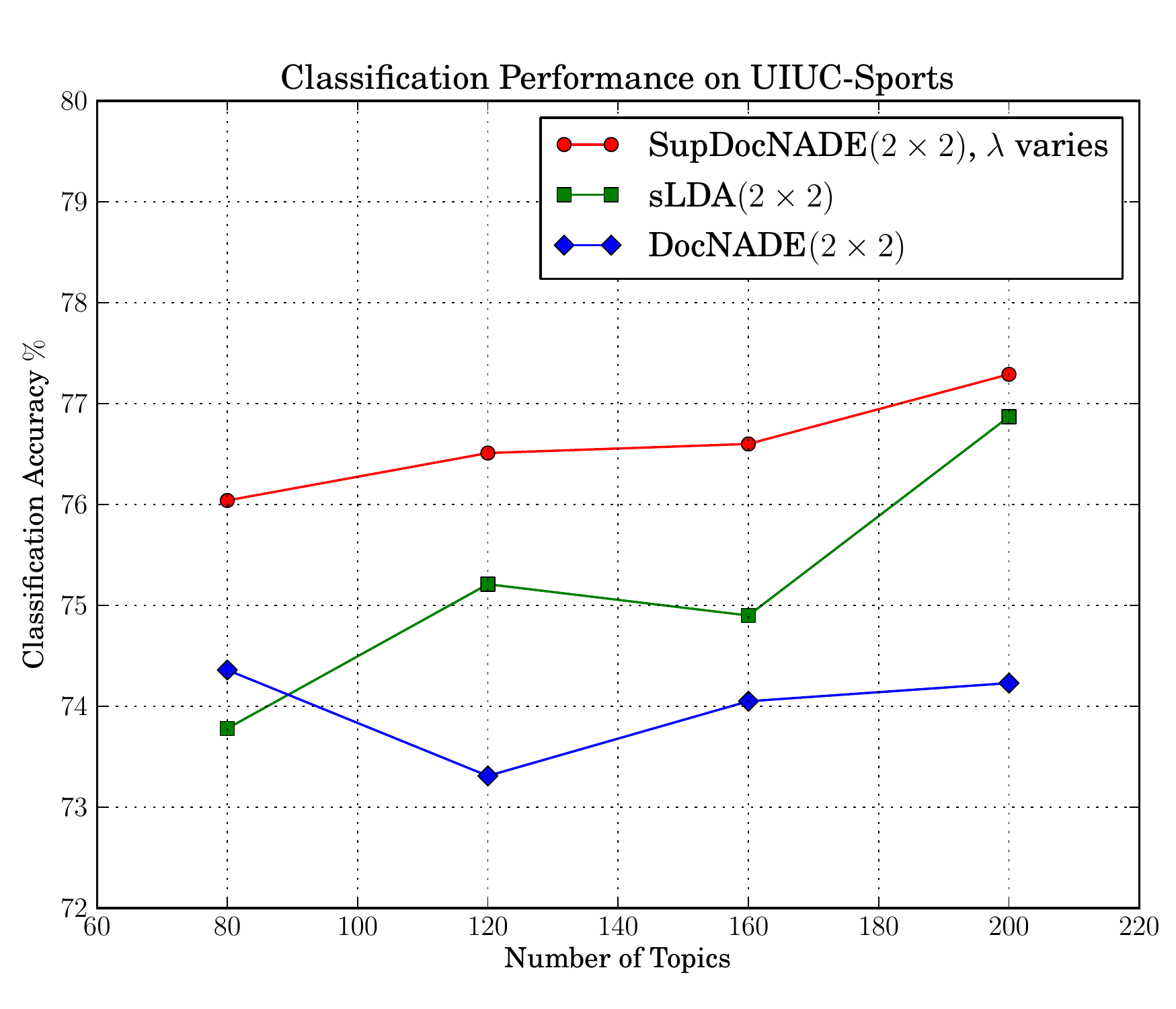}
  \includegraphics[width=0.24\linewidth]{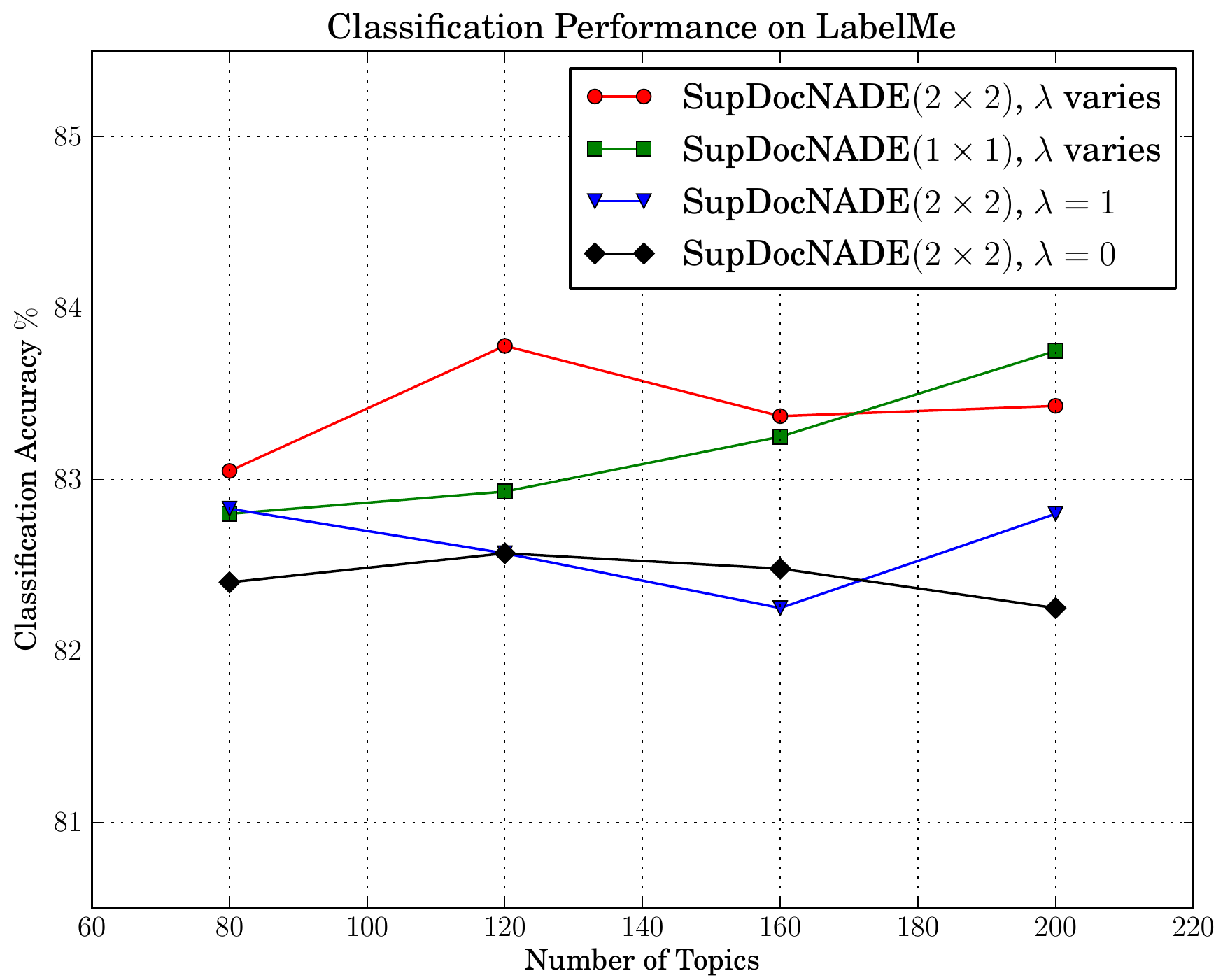}
  \includegraphics[width=0.24\linewidth]{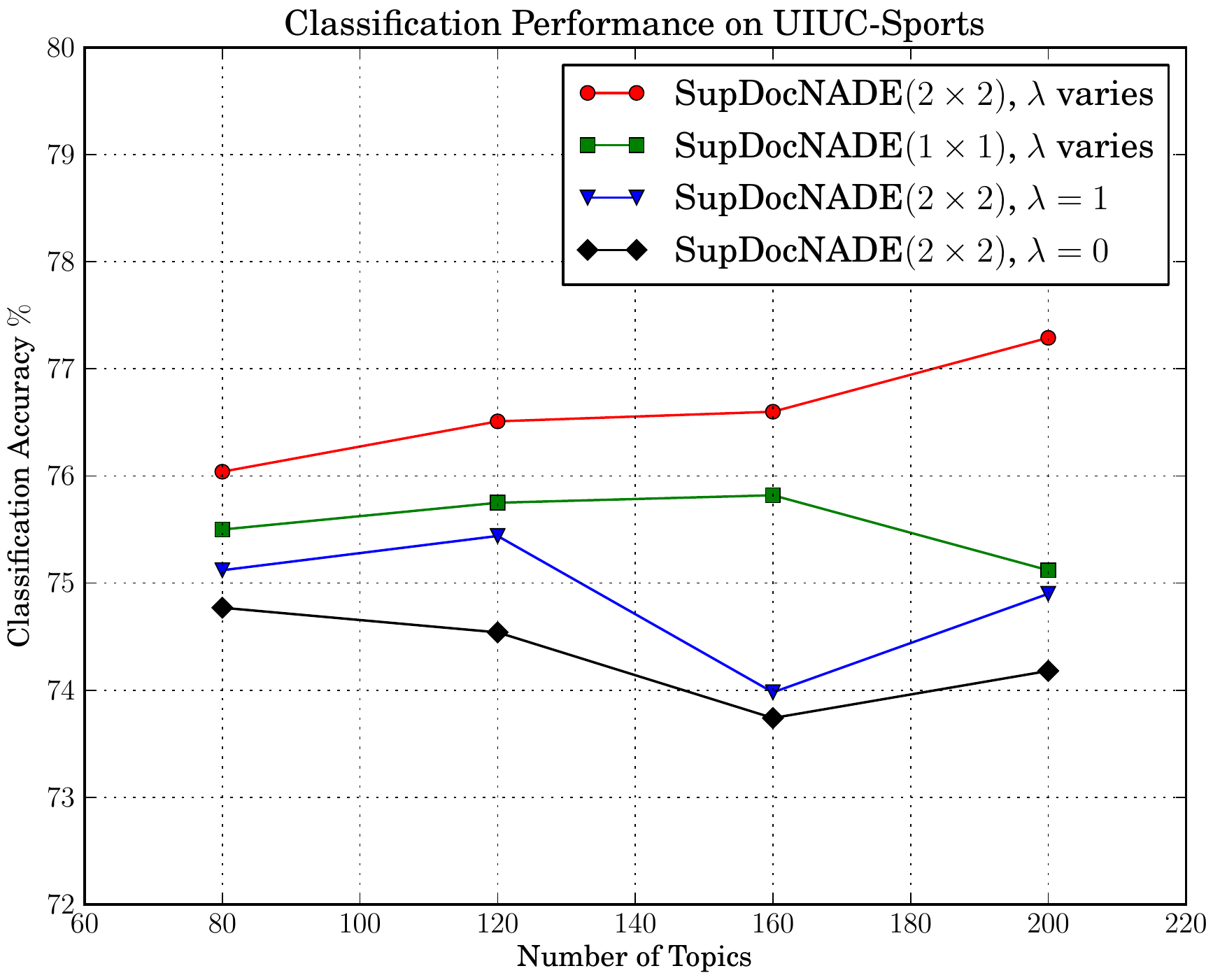}
\caption{Classification performance comparison on LabelMe (even) and UIUC-Sports (odd). On the left, we compare the classification performance of SupDocNADE, DocNADE and sLDA. On the right, we compare the performance between different variants of SupDocNADE. The ``{\it $\lambda$ varies}''  means the unsupervised weight $\lambda$ in Equation
\ref{eqn:objectfunc_hybrid} is chosen by cross-validation.}
\label{fig:labelme_and_uiuc_comp}
\end{figure*}

\begin{table}[t]
\caption{Performance comparison of SupDocNADE with different models on LabelMe and UIUC-Sports data sets.}
\begin{center}
\begin{tabular}{@{}l|l@{\hspace{1mm}}l@{\hspace{1mm}}l@{\hspace{1mm}}l@{}}
\toprule
& \multicolumn{2}{c}{LabelMe} & \multicolumn{2}{@{}c@{}}{UIUC-Sports} \\ 
\multicolumn{1}{c|}{Model} & Accuracy$\%$ & F\textup{-measure}$\%$ & Accuracy$\%$& F\textup{-measure}$\%$ \\ \midrule
SPM~\cite{lazebnik2006beyond}                  &$80.88$&$43.68 $&$72.33 $&$41.78 $   \\
MMLDA~\cite{wang2011max}                       &$81.47 ^{\dagger}$&$\mathbf{46.64} ^{\dagger \ast}$&$74.65 ^{\dagger}$&$44.51 ^{\dagger}$        \\
sLDA ~\cite{wang2009simultaneous}              &$81.87 $&$38.7 ^{\dagger}$&$76.87 $&$35.0 ^{\dagger}$       \\
DocNADE                                        &$81.97 $&$43.32 $&$74.23 $&$46.38 $\\ \midrule
SupDocNADE                                     &$\mathbf{83.43} $&$43.87 $&$\mathbf{77.29} $&$\mathbf{46.95} $\\ \bottomrule   
\end{tabular}
\begin{minipage}{0.48\textwidth}
{\small
$\dagger$: Taken from the original paper. \\
$\ast$: MMLDA performs classification and annotation separately and doesn't learn jointly from all 3 modalities.
}
\end{minipage}
\end{center}
\label{table:comparison}
\end{table}

The classification results are illustrated in
Figure~\ref{fig:labelme_and_uiuc_comp}. Similarly, we
observe that SupDocNADE outperforms DocNADE and sLDA. Tuning the
trade-off between generative and discriminative learning and
exploiting position information is usually beneficial. There
is just one exception, on LabelMe, with 200 hidden topic units, where
using a $1\times 1$ grid slightly outperforms a $2\times 2$ grid.

As for image annotation, we computed the performance
of our model with 200 topics. As shown in Table~\ref{table:comparison}, SupDocNADE obtains an
$F$\textup{-measure} of $43.87\%$  
and $46.95\%$ 
on the LabelMe and UIUC-Sports data sets respectively. This is
slightly superior to regular DocNADE.
Since code for performing image annotation using sLDA 
is not publicly available, we compare directly with the results found
in the corresponding paper~\cite{wang2009simultaneous}. \citet{wang2009simultaneous}
report $F$\textup{-measures} of $38.7\%$ and $35.0\%$ for sLDA,
which is below SupDocNADE by a large margin. 

We also compare with MMLDA~\cite{wang2011max}, which has been applied to image classification and
annotation separately. The reported classification accuracy for MMLDA
is less than SupDocNADE as shown in Table~\ref{table:comparison}.
The performance for annotation reported in ~\citet{wang2011max} is better than SupDocNADE on LabelMe
but worse on UIUC-Sports. We highlight that MMLDA did not deal with 
the class label and annotation word modalities {\it jointly}, the different modalities being
treated separately.

The spatial pyramid approach of \citet{lazebnik2006beyond} could also be adapted to perform both image 
classification and annotation. We used the code from \citet{lazebnik2006beyond} to generate 
two-layer SPM representations with a vocabulary size of 240, which is the same configuration as used by the other models. For image classification, 
an SVM with Histogram Intersection Kernel (HIK) is adopted as the classifier, as in \citet{lazebnik2006beyond}.
 For annotation, we used a $k$ nearest neighbor (KNN) prediction of the annotation words for the test images. Specifically, 
the top 5 most frequent annotation words among the $k$ nearest images (based on the SPM representation with 
HIK similarity) in the training set were selected as 
the prediction of a test image's annotation words. The number $k$ was selected by cross validation, 
for each of the 5 random splits.
 As shown in Table~\ref{table:comparison}, SPM achieves a classification accuracy of 
 $80.88\%$ and $72.33\%$ for LabelMe and UIUC-Sports, which is lower than SupDocNADE.
 As for annotation, the $F$\textup{-measure} of SPM is also lower than SupDocNADE, 
 with $43.68\%$ and $41.78\%$ for LabelMe and UIUC-Sports, respectively.

\begin{figure}[t]
\begin{center}
	\includegraphics[width=.75\linewidth]{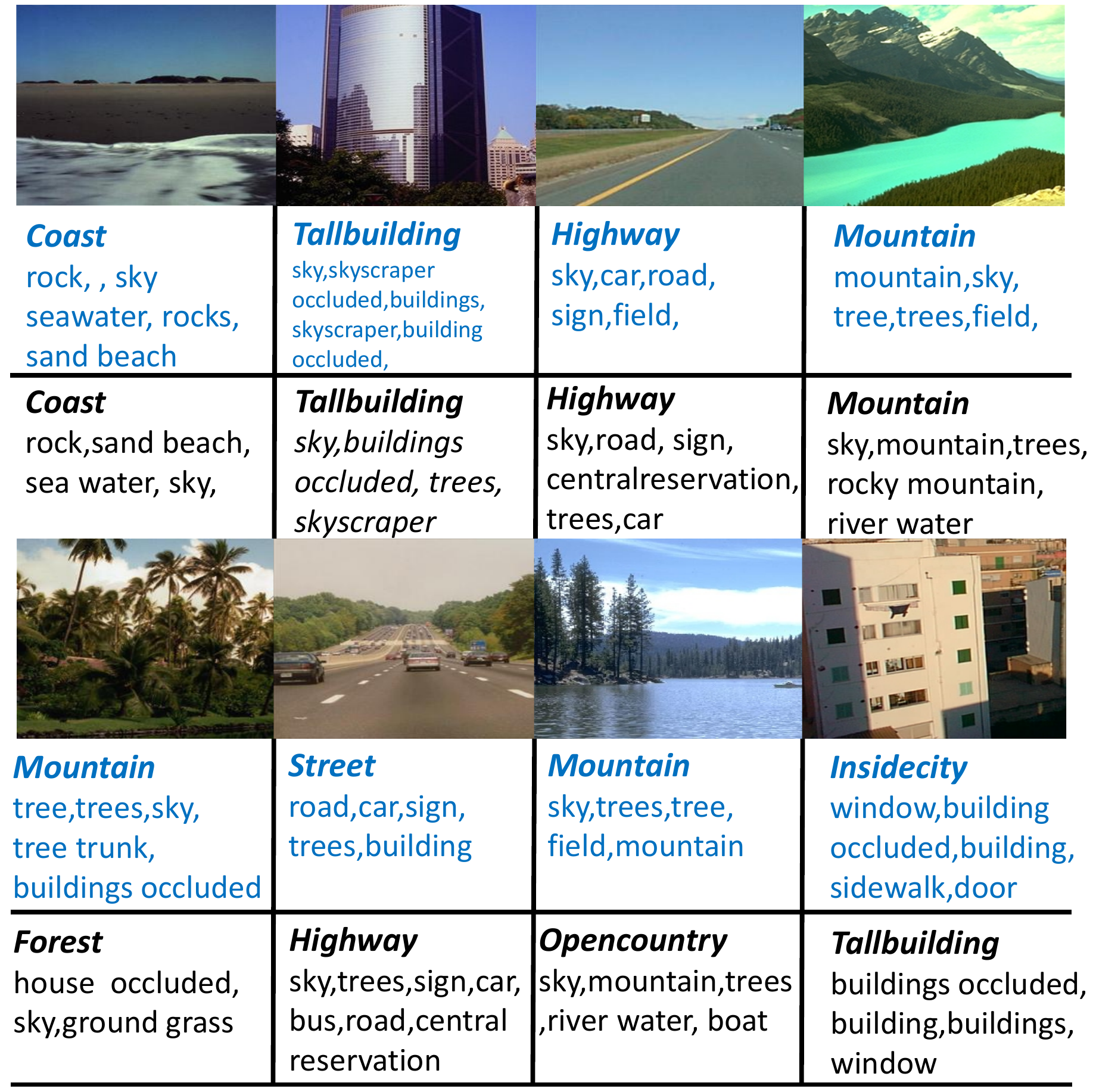}
\end{center}
\caption{Predicted class and annotation by SupDocNADE on LabelMe
   data set. We list some correctly (top row) and incorrectly (bottom
   row) classified images. The predicted (in blue) and ground-truth (in
   black) class labels and annotation words are presented under each image.}
\label{fig:figure_results}
\end{figure}

Figure~\ref{fig:figure_results} illustrates examples of correct and incorrect 
predictions made by SupDocNADE on the LabelMe data set. 


\begin{figure}[t]
\begin{center}
	\includegraphics[width=0.75\linewidth]{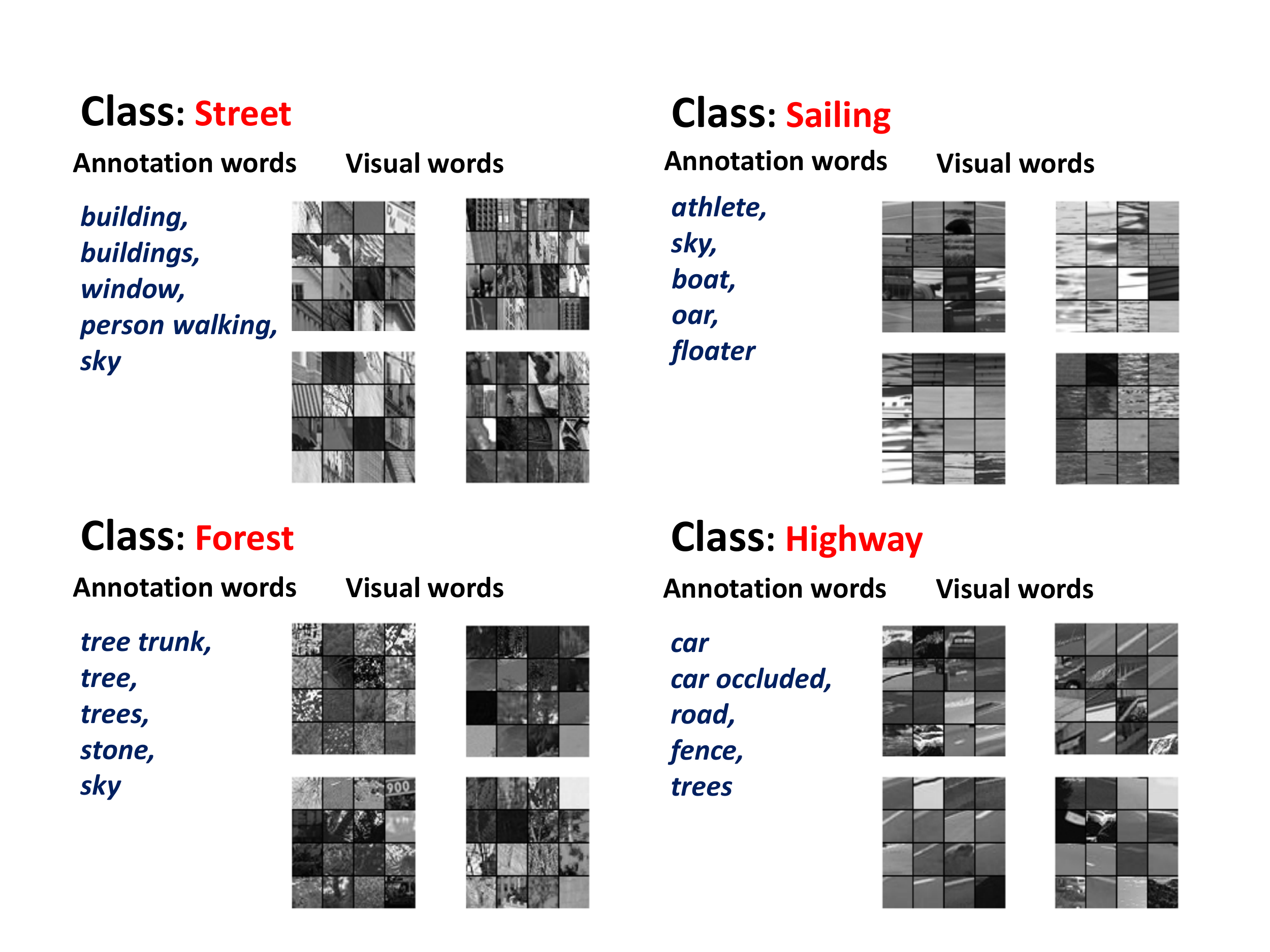}
\end{center}
\caption{Visualization of learned representations. Class labels are
  colored in red. For each class, we list 4 visual words (each represented by
  16 image patches) and 5
  annotation words that are strongly associated with each class. See
  Sec.~\ref{sec:visualization of representation} for more
  details. }
\label{fig:label_visual_anno}
\end{figure}

\subsubsection{Visualization of Learned Representations}
\label{sec:visualization of representation}
Since topic models are often used to interpret and explore the semantic structure
of image data, we looked at how we could observe the structure learned
by SupDocNADE. 

We extracted the visual/annotation words that were
most strongly associated with certain class labels within
SupDocNADE as follows. Given a class label {\it street}, which
corresponds to a column $\mathbf{U}_{:,i}$ in matrix $\mathbf{U}$, we
selected the top 3 topics (hidden units) having the largest connection
weight in $\mathbf{U}_{:,i}$. Then, we averaged the columns of matrix
$\mathbf{W}$ corresponding to these 3 hidden topics and selected the
visual/annotation words with largest averaged weight connection.
The results of this procedure for classes {\it street}, {\it sailing},
{\it forest} and {\it highway} is illustrated in Figure~\ref{fig:label_visual_anno}.
To visualize the visual words, we show 16
image patches belonging to each visual word's cluster, as extracted
by $K$-means. 
The learned associations are intuitive: for example, the
class {\it street} is associated with the annotation words ``{\it building}'',
``{\it buildings}'', ``{\it window}'', ``{\it person walking}'' and ``{\it sky}'', while the visual words
showcase parts of buildings and windows.


%




\subsection{Experiments for SupDeepDocNADE}

We now test the performance of SupDeepDocNADE, the deep extension of SupDocNADE, on the large-scale MIR Flickr data set~\cite{HuiskesM2008}. 
MIR Flickr is a challenging benchmark for multimodal data modeling task.
In this section, we will show that SupDeepDocNADE achieves state-of-the-art performance on the MIR Flickr data set over strong baselines : the DBM apporach of~\citet{srivastava2013discriminative}, MDRNN~\cite{sohn2014improved},  TagProp~\cite{guillaumin2010multimodal} and the multiple kernel learning approach of \citet{verbeek2010image}.


\subsubsection{MIR Flickr Data Set}
\label{sec:MIR intro}
The MIR Flickr data set contains $1$ million real images that are collected from the image hosting website Flickr. The social tags of each image are also collected and used as annotations in our experiments.
Among the $1$ million images, there are $25~000$ images that is labeled into $38$ classes, such as {\it sky, bird, people, animals, car, etc.}, giving us a subset of labeled images. Each image in the labeled subset can have multiple class labels. In our experiments, we used $15~000$ images for training and $10~000$ images for testing. The remaining $975~000$ images do not have labels and thus were used for the unsupervised pretraining of SupDeepDocNADE (see next section). The most frequent $2000$ tags are collected for the 
 annotation vocabulary, following previous work~\cite{srivastava2012multimodal,srivastava2013discriminative}. The averaged number of annotations for an image is $5.15$. In the whole
 data set, $128~501$ images do not have annotations, out of which $4551$ images are in the labeled subset.

\subsubsection{Experimental Setup for SupDeepDocNADE}
\label{sec:SupDeepDocNADE config}

In order to compare directly with the DBM approach of \citet{srivastava2013discriminative}, we use the same experimental configuration. Specifically, the images in MIR Flickr are first 
rescaled to make the maximum side of each image be $480$ pixels, keeping the aspect ratio. Then, $128$ dimensional SIFT features are densely
sampled on these images to extract the visual words. Following \citet{srivastava2013discriminative}, we used $4$ different scales of patch size, which are $4, 6, 8, 10$ pixels, respectively, and the patch step is fixed to $3$ pixels. The SIFT features from the unlabeled images were quantized into 2000 clusters, which is used as the visual word vocabulary. Thus, the image modality  is represented by the bag of visual words representation using this vocabulary. 
As preliminary experiments suggested that spatial information (see Section~\ref{sec:multiple regions}) wasn't useful on the Flickr data set, we opted for not using it here.
Similarly, the text modality for SupDeepDocNADE is represented using the annotation vocabulary, which is built upon the most frequent
 $2000$ tags, as is mentioned in Section~\ref{sec:MIR intro}. The visual words and annotation words are combined together and treated as the input of SupDeepDocNADE.

As for the global features (Section~\ref{sec:global_features}), a combination of Gist~\cite{oliva2001modeling} and MPEG-7 descriptors~\cite{manjunath2001color}(EHD, HTD, CSD, CLD, SCD) is adopted in our experiments, as in \citet{srivastava2013discriminative}. The length of the global features is $1857$. 

We used a $3$ hidden layers architecture in our experiments, with the size of each hidden layer being $2048$.  Note that the DBM~\cite{srivastava2012multimodal,srivastava2013discriminative} also use $3$ hidden layers with $2048$ hidden units for each layer, thus our comparison with the DBM is fair.  The activation function for the 
hidden units is the rectified linear function. We used a softmax output layer instead of a binary tree to compute the conditionals $p\left(v_{o_{d}}|\mathbf{v}_{o_{<d}}, \theta,o_{<d},o_{d}\right)$ for SupDeepDocNADE, as discussed in Section~\ref{sec: DeepDocNADE}.

For the prediction of class labels, since images in MIR Flickr could have multiple labels, we used a sigmoid output layer instead of the softmax to compute the probability that an image
belongs to a specific class $c_i$
\begin{equation}
p\left(c_i=1|\mathbf{v},\mathbf{\theta}\right)= {\rm sigmoid}\left(d_{c_i}+\mathbf{U}_{c_i,:}\mathbf{h}^{\left(N\right)}\right)
\end{equation}
where $\mathbf{h}^{\left(N\right)}$ is the hidden representation of the top layer. As a result, the supervised cost part in Equation~\ref{eqn:SupdeepDocNADE_hybrid} is replaced by the cross entropy $\sum_{i=1}^{C} - c_i\log p\left(c_i=1|\mathbf{v},\mathbf{\theta}\right)-\left(1- c_i\right)\log p\left(c_i=0|\mathbf{v},\mathbf{\theta}\right)$, where $C$ is the number of classes.
 
In all experiments, the unlabeled images are used for unsupervised pretraining. This is achieved by first training a DeepDocNADE model, without any output layer predicting class labels. The result of this training is then used to initialize the parameters of a SupDeepDocNADE model, which is finetuned on the labeled training set based on the loss of Equation~\ref{eqn:SupdeepDocNADE_hybri_weight}. 


Once training is finalized, the hidden representation from the top hidden layer after observing all words (both visual words and annotation words) of an image is feed
 to a linear SVM~\cite{fan2008liblinear} to compute confidences of an image belonging to each class. The average precision (AP) for each class is obtained based on these confidences, where AP is the area under the precision-recall curve. After that, the mean average precision (MAP) over all classes is computed and used as the metric to
 measure the performance of the model.
We used the same $5$ training/validation/test set splits on the labeled subset of MIR Flickr as \citet{srivastava2013discriminative} and report the average performance on the $5$ splits.

To initialize the connection matrices, we followed the recommendation of \citet{glorot2010understanding} used a uniform distribution:
\begin{equation}
\Theta \sim U\left[-\frac{\sqrt{6}}{\sqrt{l_{\Theta}+w_{\Theta}}}, \frac{\sqrt{6}}{\sqrt{l_{\Theta}+w_{\Theta}}}\right]
\end{equation}
where $\Theta \in \lbrace \mathbf{W}, \mathbf{U}, \mathbf{V} \rbrace$ is a connection matrix, $l_{\Theta}$,$w_{\Theta}$ are the number of rows and columns respectively of matrix $\Theta$, respectively, and $U$ is the uniform distribution.
In practice, we've also found it useful to normalize the input histograms $\mathbf{\tilde{x}}\left({\bf v}_{o_{<d}}\right)$ for each image, by rescaling them to have unit variance.

The hyper-parameters (learning rate, unsupervised weight $\lambda$, and the parameter for linear SVM, etc.) are chosen by cross-validation. 
To prevent overfitting, dropout~\cite{hinton2012improving} is adopted during training, with a dropout rate of $0.5$ for all hidden layers. 
We also maintained an exponentially decaying average of the 
parameter values throughout the gradient decent training procedure and used the averaged parameters at test time. This corresponds to 
Polyak averaging~\citep{swersky2010tutorial}, but where the linear average is replaced by a weighting that puts more emphasis on recent parameter values.
For the annotation weight, it was fixed to $12~000$, which is approximately the ratio 
of the averaged visual words and annotation words of the data set. We will investigate the impact of the annotation weight on the performance in Section~\ref{sec:SupDeepDocNADE annoweight}.

\begin{table}[t]
\caption{Performance comparison on MIR Flickr data set.}
\begin{center}
\begin{tabular}{@{}l|l@{\hspace{1mm}}l@{\hspace{1mm}}l@{\hspace{1mm}}l@{}}
\toprule
 
\multicolumn{1}{c|}{Model} & MAP\\ \midrule
TF-IDF &$0.384 \pm 0.004$ \\
Multiple Kernel Learning SVMs~\cite{guillaumin2010multimodal}                  &$0.623$ \\
TagProp~\cite{verbeek2010image}                       &$0.640$        \\
Multimodal DBM~\cite{srivastava2013discriminative}              &$0.651\pm 0.005$      \\
MDRNN~\cite{sohn2014improved}              &$0.686\pm 0.003$      \\ \midrule
SupDeepDocNADE (1 hidden layer, 625 epochs pretraining)                                   &$0.654\pm 0.004$\\ 
SupDeepDocNADE (2 hidden layers, 625 epochs pretraining)                                   &$0.671\pm 0.006$\\  
SupDeepDocNADE (3 hidden layers, 625 epochs pretraining)                                   &$0.670\pm 0.005$\\ \midrule
SupDeepDocNADE (2 hidden layers, 2325 epochs pretraining)                                   &$0.682\pm 0.005$\\
SupDeepDocNADE (3 hidden layers, 2325 epochs pretraining)                                   &$0.686\pm 0.005$\\ \midrule
SupDeepDocNADE (2 hidden layers, 4125 epochs pretraining)                                   &$0.684\pm 0.005$\\
SupDeepDocNADE (3 hidden layers, 4125 epochs pretraining)                                   &$\mathbf{0.691}\pm 0.005$\\ \bottomrule 
\end{tabular}
\begin{minipage}{0.48\textwidth}

\end{minipage}
\end{center}
\label{table:comparison_DBM}
\end{table}

\begin{figure}[h]
\begin{center}
	\includegraphics[width=0.7\linewidth]{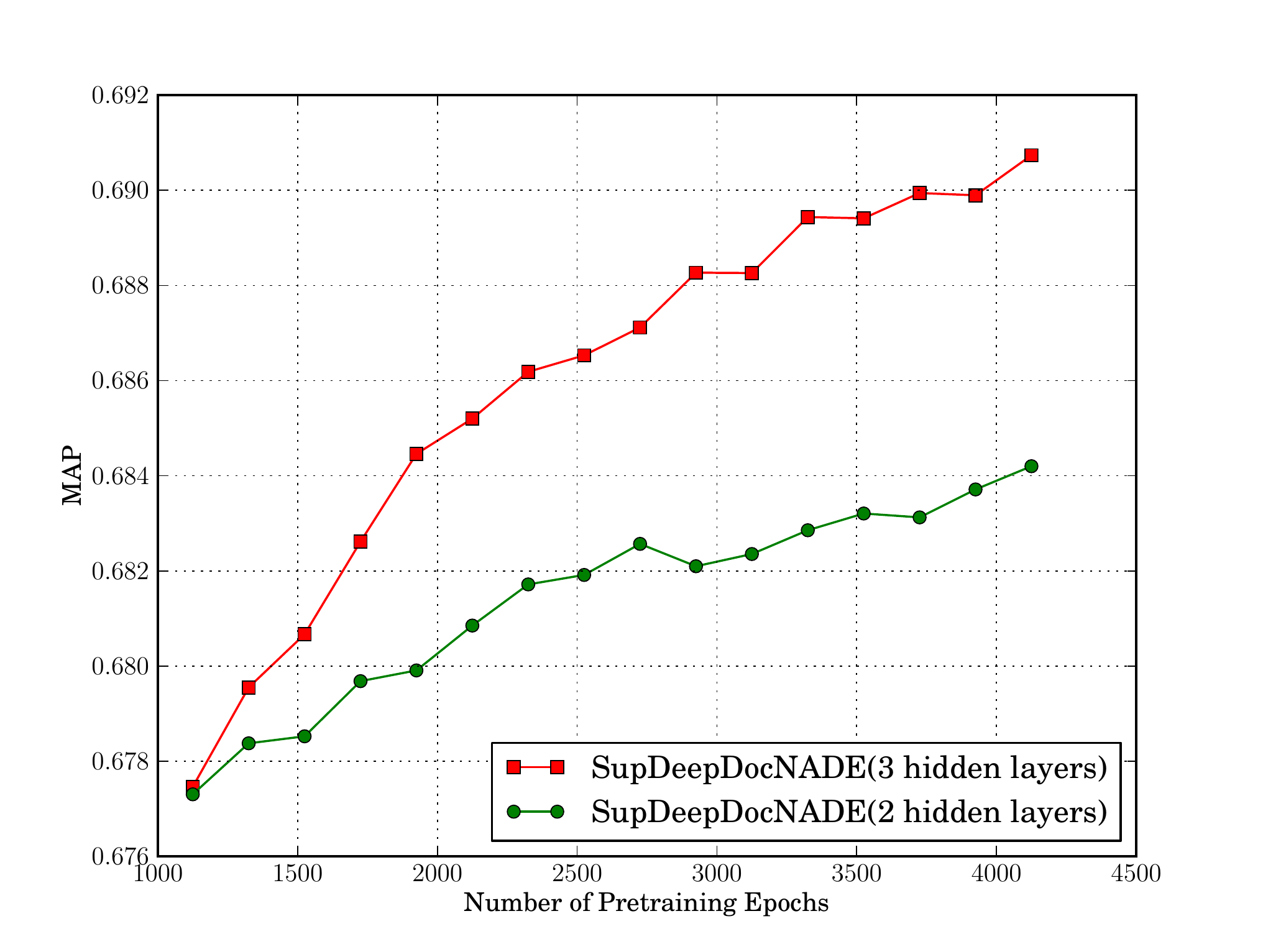}
\end{center}
\caption{Performance of SupDeepDocNADE w.r.t the number of epochs pretrained on unlabeled data.
}
\label{fig:map_pretrain}
\end{figure}

\subsubsection{Comparison with other baselines}
\label{sec:compare with DBM}
Table~\ref{table:comparison_DBM} presents a comparison of the performance of SupDeepDocNADE with the DBM approach of \citet{srivastava2013discriminative} and MDRNN of \citet{sohn2014improved} as well as other strong baselines, in terms of MAP performance. We also provide the simple and popular TF-IDF baseline in Table~\ref{table:comparison_DBM} to make the comparison more complete. The TF-IDF baseline is conducted only on the bag-of-words representations of images without global features. We feed the TF-IDF representations to a linear SVM to obtain confidences of an image belonging to each class and then we compute the Mean AP, as for SupDeepDocNADE. 

We can see that SupDeepDocNADE achieves the best performance among all  methods. 
More specifically, we first pretrained the model for $625$ epochs on the unlabeled data with $1$, $2$ and $3$ hidden layers. The results illustrated in Table~\ref{table:comparison_DBM} show that SupDeepDocNADE outperforms the DBM baseline by a large margin. Moreover, we can see that SupDeepDocNADE with $2$ and $3$ hidden layers performs better than with only $1$ hidden layer, with $625$ epochs of pretraining. We then pretrained the model for more epochs on the unlabeled data ($2325$ epochs). As shown in Table~\ref{table:comparison_DBM}, with more pretraining epochs, the deeper model ($3$ hidden layers) performs even better. This confirms that the use of a deep architecture is beneficial. When the number of pretraining epochs reaches $4125$, the SupDeepDocNADE model with $3$ hidden layers achieves a MAP of $0.691$, which outperforms all the strong baselines and increases the performance gap  with the 2-hidden-layers model. 

From Tabel~\ref{table:comparison_DBM} we can also see that the performance of 2-layers SupDeepDocNADE does not improve as much as 3-layers SupDeepDocNADE when the number of pretraining epochs increases from $2325$ to $4125$. Figure~\ref{fig:map_pretrain} shows the the performance of SupDeepDocNADE w.r.t the number of pretraining epochs. We can see from Figure~\ref{fig:map_pretrain} that with more epochs of pretraining, the performance of 3-layers SupDeepDocNADE increases faster than the 2-layers models, which indicates that the capacity of 3-layers SupDeepDocNADE is bigger than the 2-layers model and the capacity could be leveraged by more pretraining. Figure~\ref{fig:map_pretrain} also suggests that the performance of SupDeepDocNADE could be even better than $0.691$ with more pretraining epochs.

\begin{figure}[t]
\begin{center}
  \includegraphics[width=0.65\linewidth]{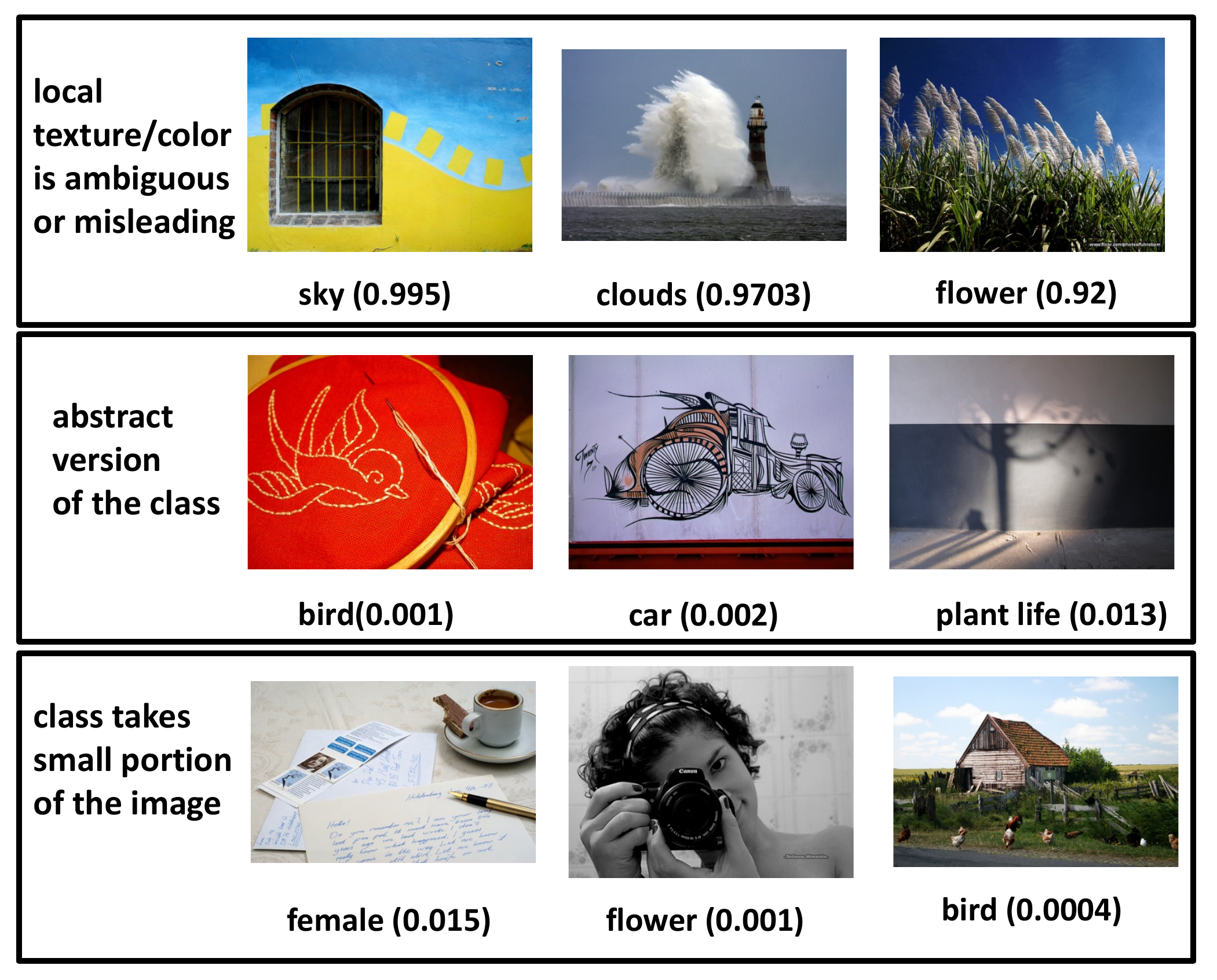}
\end{center}
\caption{Illustration of some failed examples of SupDeepDocNADE. The reasons for failure are listed on the left-side of each row. For each reason, 
we list $3$ examples. The text below each image is the confidence of either the wrongly predicted class (the top row) or the ground truth class (the middle and bottom rows). The maximum value of confidence is $1$ and minimum is $0$.}
\label{fig:failure_analysis}
\end{figure}

Figure~\ref{fig:failure_analysis} illustrates some failed predictions of SupDeepDocNADE, where the reasons for failure are shown on the left-side of each row. One of the reasons for failure is that the local texture/color is ambiguous or misleading. For example, in the first image of the top row, the blue color in the upper side of the wall misleads the model to predict "sky" with a confidence of $0.995$. Another type of failure, which is shown in the middle row of Figure~\ref{fig:failure_analysis}, is caused by images of an abstract illustration of the class. For instance, the model fails to recognize the bird, car and tree in the images of the middle row, respectively, as these  images are merely abstract illustrations of these concepts. The third reason illustrated in the bottom row is that the class takes a small portion of the image, making it more likely to be ignored. For example, the female face on the stamp in the first image of the bottom row is too small to be recognized by the model. Note that we just illustrated some failed examples and there might be other kinds of failures. In practice, we also 
find that some images are not correctly labeled, which might also cause some failures.

Having established that SupDeepDocNADE achieves state-of-the-art performance on the MIR Flickr data set  and also discussed some failed examples, we now explore in more details some of its properties in the
following sections.

\begin{figure}[t]
\begin{center}
	\includegraphics[width=0.6\linewidth]{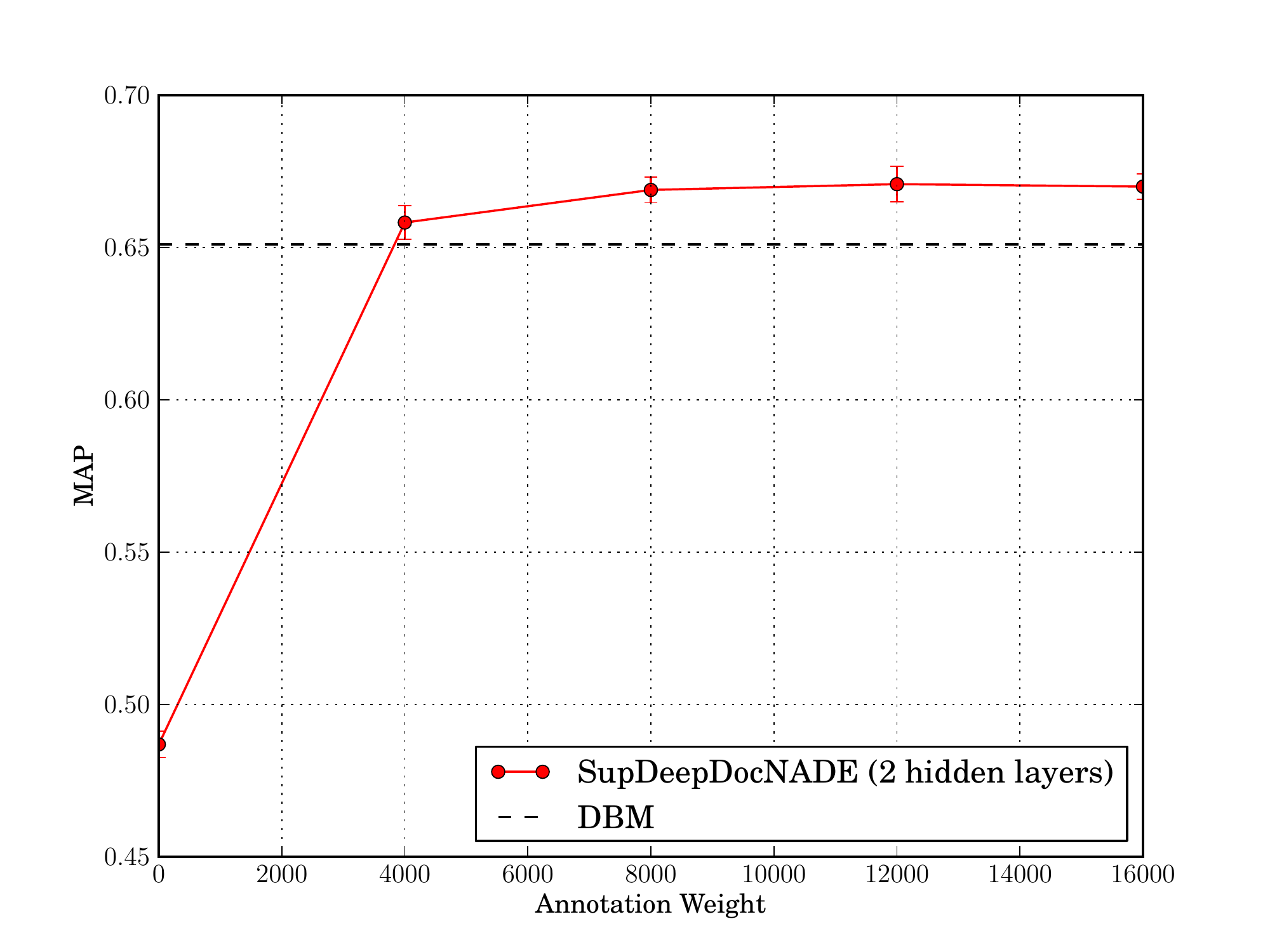}
\end{center}
\caption{Comparison between different annotation weights. 
}
\label{fig:anno_weight}
\end{figure}

\begin{figure}[t]
\begin{center}
  \includegraphics[width=0.7\linewidth]{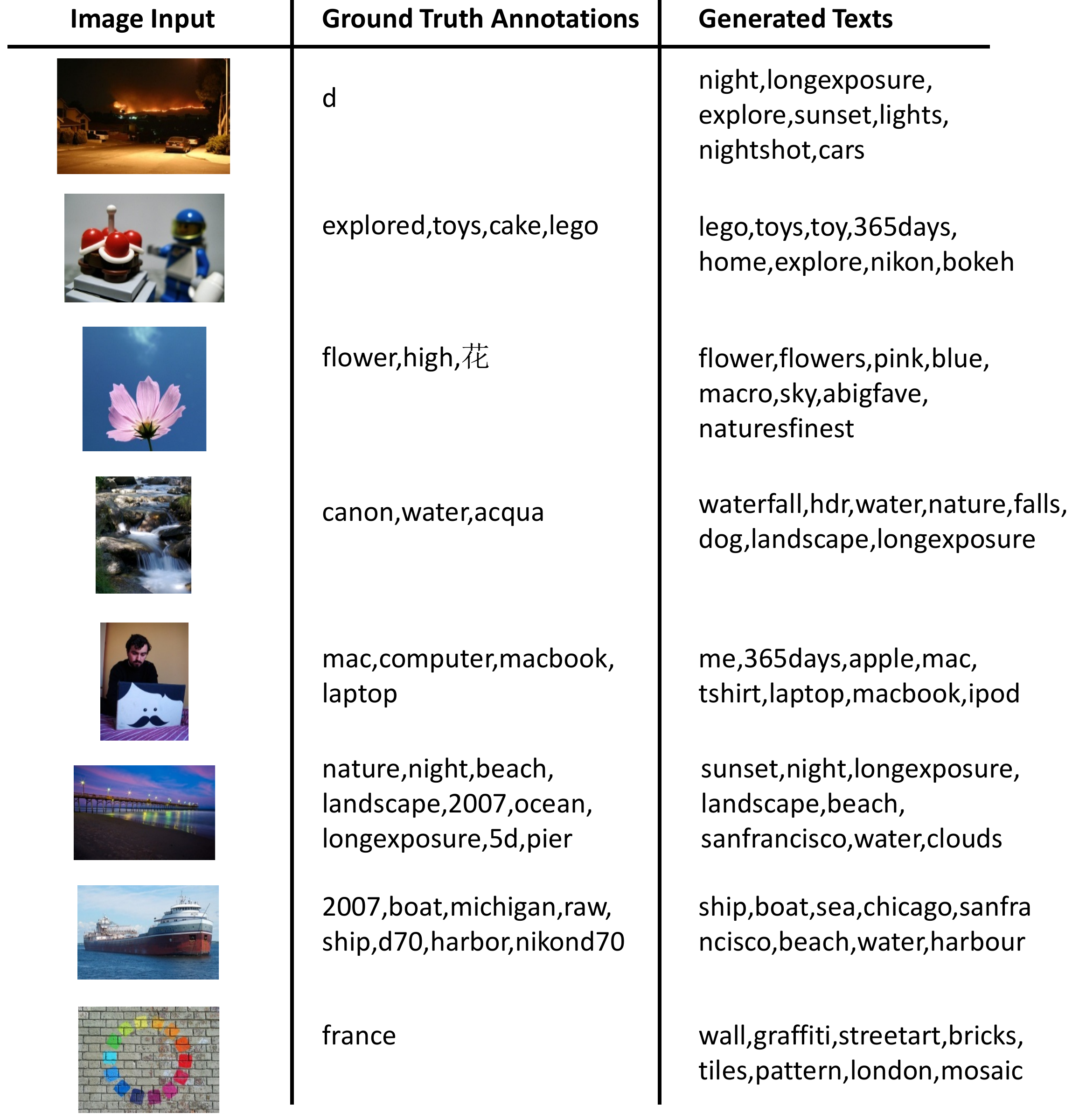}
\end{center}
\caption{The illustration of generated texts from images by SupDeepDocNADE. The input for this task is the image modality only and the output is the generated text. We put the ground truth annotations in the second column and illustrate the top $8$ words generated using SupDeepDocNADE in the third column. If there is no ground truth annotations, the corresponding part is left blank. We can see that SupDeepDocNADE can generate meaningful annotations from images.}
\label{fig:predict_anno}
\end{figure}

\begin{figure}[t]
\begin{center}
	\includegraphics[width=0.9\linewidth]{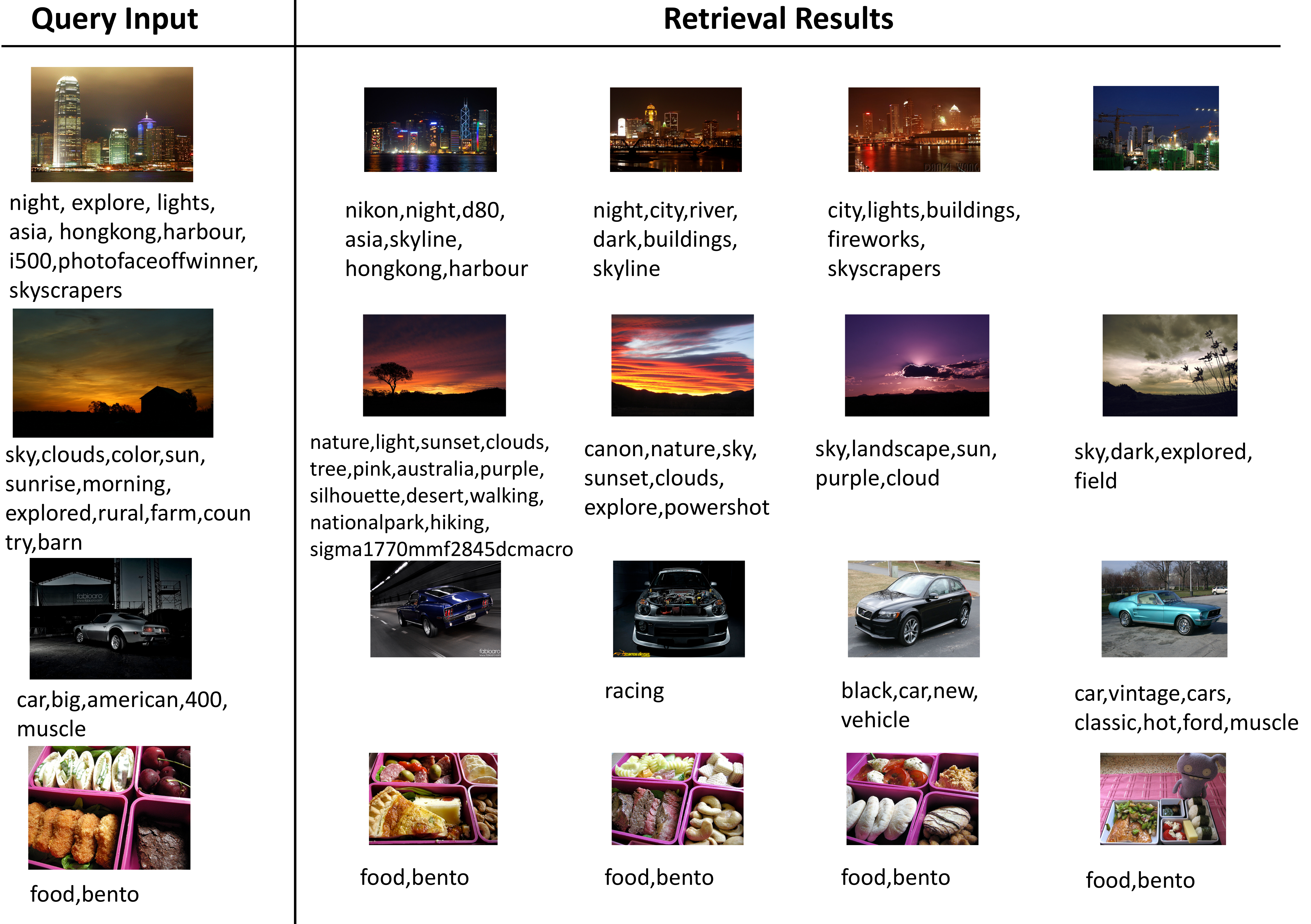}
\end{center}
\caption{The illustration of multimodal retrieval results for SupDeepDocNADE. 
Both the query input and retrieved results contain image and text modalities. The annotations (text modality) are shown 
under the image. The query input is shown in the first column, and the $4$ most similar image/annotation pairs according to SupDeepDocNADE are shown in the following columns, ranked by similarity from left to right.  }
\label{fig:multimodal_retrieve}
\end{figure}

\subsubsection{The Impact of the Annotation Weight}
\label{sec:SupDeepDocNADE annoweight}

In Section~\ref{sec:SupDeepDocNADE annoweight}, we proposed to weight differently the annotation words to deal with the problem of imbalance in the number of visual and annotation words. In this part, we investigate the influence of the annotation weight on the performance. Specifically, we set the annotation weight to $\left\lbrace 1,4000,8000,12~000,16~000 \right\rbrace$, and
show the performance for each of the annotation weight values. Note that when the annotation weight equals $1$, there is no compensation for the imbalance of visual words and annotation words.
The other experimental configurations are the same as in Section~\ref{sec:SupDeepDocNADE config}. 

Figure~\ref{fig:anno_weight} shows the performance comparison between different annotation weights. As expected, SupDeepDocNADE performs extremely bad when the annotation equals to $1$, 
When the annotation weight is increased, the performance gets better. Among all the chosen annotation weights, $12~000$ performs best, which achieves a MAP of $0.671$. The other annotation weights also achieves good performance compared with the DBM model~\citep{srivastava2013discriminative}: MAP of $0.658$, $0.669$ and $0.670$ for annotation weight values of $4000$, $8000$ and $16~000$, respectively.

\subsubsection{Visualization of the Retrieval Results}
\label{sec:SupDeepDocNADE demo}
Since SupDeepDocNADE is used for multimodal data modeling, we illustrate here some results for multimodal data retrieval tasks.
More specifically, we show some qualitative results in two multimodal data retrieval scenarios: multimodal data query and generation of text from images.\\
\textbf{Multimodal Data Query:}  Given a query corresponding to an image/annotation pair,  the task is to retrieve other similar pairs from a collection, using the hidden representation learned by SupDeepDocNADE. In this task, the cosine similarity is adopted as the 
similarity metric. In this experiment, each query corresponds to an individual test example and the collection corresponds to the rest of the test set.
Figure~\ref{fig:multimodal_retrieve} illustrates the retrieval results for multimodal data query task, where we show the $4$ most similar images to the query input in the 
testset.\\
\textbf{Generating Text from Image:} As SupDeepDocNADE learns the relationship between the image and text modalities, we test its ability to generate
text from given images. This task is implemented by feeding SupDeepDocNADE  only the bag of visual words and selecting the annotation words according to their probability of being the next word, similarly to Section~\ref{sec: anno}. Figure~\ref{fig:predict_anno} illustrates the ground truth annotation and the most probable $8$ annotations generated by SupDeepDocNADE. We can see that SupDeepDocNADE generated very meaningful texts according to the image modality, which shows that it effectively learned about the statistical structure between the two modalities.

\section{Conclusion and Discussion}
\label{conclusion}

In this paper, we proposed SupDocNADE, a supervised extension of
DocNADE, which can learn jointly from visual words,
annotations and class labels. Moreover, we proposed a deep extension of SupDocNADE which outperforms its shallow version and can be trained efficiently.
 Although both SupDocNADE and SupDeepDocNADE are the same in nature, SupDeepDocNADE  differs from the single layer version in its training process. Specifically, the training process of SupDeepDocNADE is performed over a subset of the words by summing the gradients over several orderings sharing the same permutation up to a randomly selected position $d$, while the single layer version does the opposite and exploits a single randomly selected ordering but updates all the conditionals on the words.

Like all topic models, our model is trained to model the
distribution of the bag-of-word representations of images and can
extract a meaningful representation from it. Unlike most topic models
however, the image representation is not modeled as a latent random
variable in a model, but instead as the hidden layer of a neural autoregressive
network. A distinctive advantage of SupDocNADE is that it does not require
any iterative, approximate inference procedure to compute an image's
representation. Our experiments confirm that SupDocNADE is a
competitive approach for multimodal data modeling and SupDeepDocNADE achieves state-of-the-art performance on the 
challenging multimodal data benchmark MIR Flickr.

{\small
\bibliographystyle{IEEEtranNAT}
\bibliography{egbib}
}

\end{document}